\documentclass{article}

\PassOptionsToPackage{numbers, comma}{natbib}

\usepackage[final]{neurips_2018}

\usepackage{authblk}

\usepackage[utf8]{inputenc} 
\usepackage[T1]{fontenc}    
\usepackage{hyperref}       
\usepackage{url}            
\usepackage{booktabs}       
\usepackage{amsfonts}       
\usepackage{nicefrac}       
\usepackage{microtype}      
\usepackage{xspace}
\usepackage{exscale}       
\usepackage{amsmath,bm}

\usepackage{graphicx}
\graphicspath{ {./} } 
\usepackage{amsmath,amssymb} 
\usepackage{color}

\usepackage{dsfont}        

\newcommand{\R}{\mathds{R}}
\renewcommand{\vec}[1]{\boldsymbol{\mathbf{#1}}}
\usepackage{siunitx}
\setcitestyle{square}
\usepackage{caption}
\usepackage{subcaption}
\usepackage[title]{appendix}
\usepackage{float}

\hypersetup{colorlinks,linkcolor={blue},citecolor={blue},urlcolor={red}}  
\newcolumntype{C}[1]{>{\centering\let\newline\\\arraybackslash\hspace{0pt}}m{#1}}

\newcommand{\unet}{U-Net}
\newcommand{\probunet}{Probabilistic U-Net}
\newcommand{\ensemble}{U-Net Ensemble}
\newcommand{\dropoutunet}{Dropout U-Net}
\newcommand{\itoivae}{Image2Image VAE}
\newcommand{\mheads}{M-Heads}
\newcommand{\lungs}{lung abnormalities}

\title{A Probabilistic U-Net for Segmentation of Ambiguous Images} 

\author[1\thanks{work done during an internship at DeepMind.}\;,2,]{\textbf{Simon A. A. Kohl}}
\author[1] {\textbf{Bernardino Romera-Paredes}}
\author[1] {\textbf{Clemens Meyer}}
\author[1] {\textbf{Jeffrey De Fauw}}
\author[1] {\textbf{Joseph R. Ledsam}}
\author[2] {\textbf{Klaus H. Maier-Hein}}
\author[1] {\textbf{S. M. Ali Eslami}}
\author[1] {\textbf{Danilo Jimenez Rezende}}
\author[1] {\textbf{Olaf Ronneberger}}
\affil[1]{DeepMind, London, UK}
\affil[2]{Division of Medical Image Computing, German Cancer Research Center, Heidelberg, Germany}
\affil[ ]{\texttt{\{simon.kohl,k.maier-hein\}@dkfz.de}}
\affil[ ]{\texttt{\{brp,meyerc,defauw,jledsam,aeslami,danilor,olafr\}@google.com}}

\begin{document}

\maketitle
\begin{abstract}
  Many real-world vision problems suffer from inherent ambiguities. In clinical applications for example, it might not be clear from a CT scan alone which particular region is cancer tissue. Therefore a group of graders typically produces a set of diverse but plausible segmentations. We consider the task of learning a distribution over segmentations given an input. To this end we propose a generative segmentation model based on a combination of a \unet{} with a conditional variational autoencoder that is capable of efficiently producing an unlimited number of plausible hypotheses. We show on a \lungs{} segmentation task and on a  Cityscapes segmentation task that our model reproduces the possible segmentation variants as well as the frequencies with which they occur, doing so significantly better than published approaches. These models could have a high impact in real-world applications, such as being used as clinical decision-making algorithms accounting for multiple plausible semantic segmentation hypotheses to provide possible diagnoses and recommend further actions to resolve the present ambiguities.
\end{abstract}

\section{Introduction}
The semantic segmentation task assigns a class label to each pixel in an image. While in many cases the context in the image provides sufficient information to resolve the ambiguities in this mapping, there exists an important class of images where even the full image context is not sufficient to resolve all ambiguities. Such ambiguities are common in medical imaging applications, e.g., in \lungs{} segmentation from CT images. A lesion might be clearly visible, but the information about whether it is cancer tissue or not might not be available from this image alone. Similar ambiguities are also present in photos. E.g. a part of fur visible under the sofa might belong to a cat or a dog, but it is not possible from the image alone to resolve this ambiguity\footnote{In \cite{lee2016stochastic} this is defined as \textit{ambiguous evidence} in contrast to \textit{implicit class confusion}, that stems from an ambiguous class definition (e.g. the concepts of desk vs. table). For the presented work this differentiation is not required.}. Most existing segmentation algorithms either provide only one likely consistent hypothesis (e.g., ``all pixels belong to a cat'') or a pixel-wise probability (e.g., ``each pixel is 50\% cat and 50\% dog'').  

Especially in medical applications where a subsequent diagnosis or a treatment depends on the segmentation map, an algorithm that only provides the most likely hypothesis might lead to misdiagnoses and sub-optimal treatment. Providing only pixel-wise probabilities ignores all co-variances between the pixels, which makes a subsequent analysis much more difficult if not impossible. If multiple consistent hypotheses are provided, these can be directly propagated into the next step in a diagnosis pipeline, they can be used to suggest further diagnostic tests to resolve the ambiguities, or an expert with access to additional information can select the appropriate one(s) for the subsequent steps.

\begin{figure}[tbp]
\centering
\includegraphics[width=\textwidth]{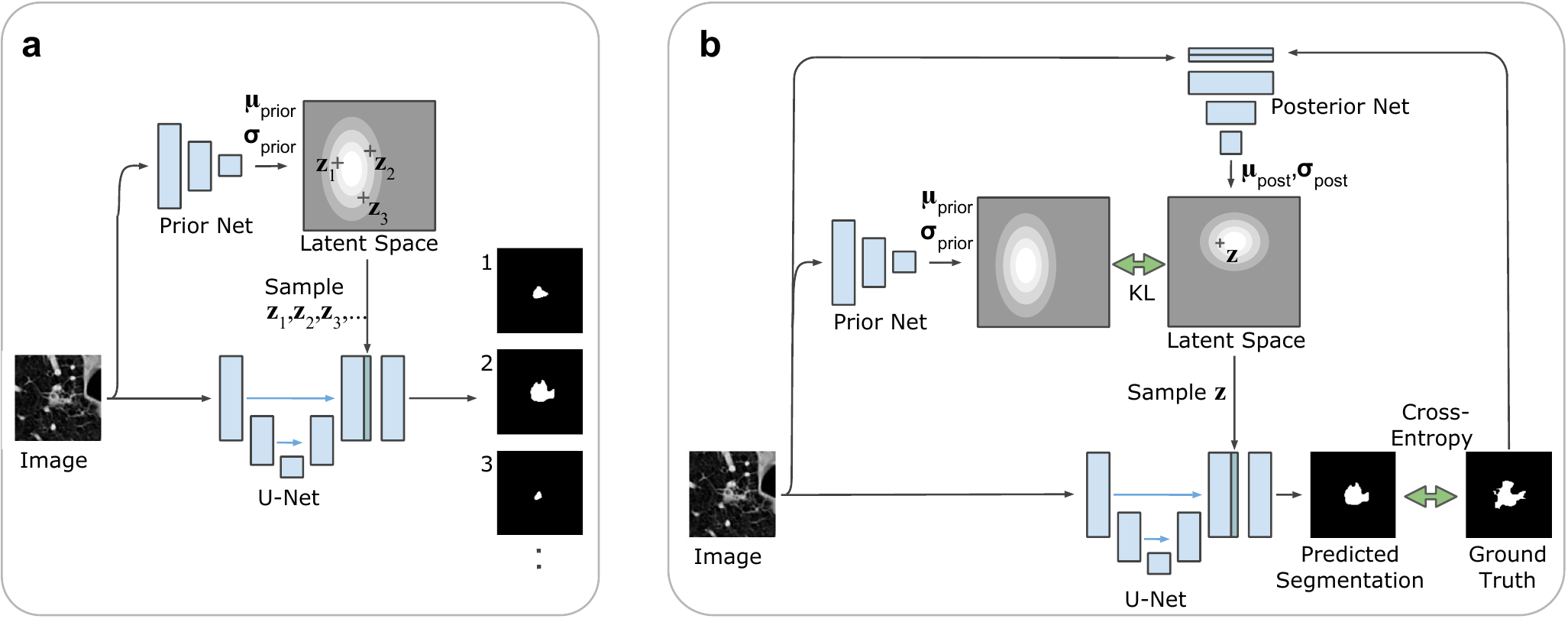}
\captionsetup{format=hang}
\caption{The \probunet.  (\textbf{a}) Sampling process. Arrows: flow of operations;  blue blocks: feature maps. The heatmap represents the probability distribution in the low-dimensional latent space $\R^N$ (e.g., $N=6$ in our experiments). For each execution of the network, one sample $\vec{z} \in \R^N$ is drawn to predict one segmentation mask. Green block: $N$-channel feature map from broadcasting sample $\vec{z}$. The number of feature map blocks shown is reduced for clarity of presentation. (\textbf{b}) Training process illustrated for one training example. Green arrows: loss functions. }
\label{fig:architecture}
\end{figure}

Here we present a segmentation framework that provides multiple segmentation hypotheses for ambiguous images (\autoref{fig:architecture}a). 
Our framework combines a conditional variational auto encoder (CVAE) \cite{vae1,vae2,vae3,vae4} which can model complex distributions, with a \unet{} \cite{Ronneberger2015} which delivers state-of-the-art segmentations in many medical application domains. A low-dimensional latent space encodes the possible segmentation variants. A random sample from this space is injected into the \unet{} to produce the corresponding segmentation map. One key feature of this architecture is the ability to model the joint probability of all pixels in the segmentation map. This results in multiple segmentation maps, where each of them provides a consistent interpretation of the whole image. Furthermore our framework is able to also learn hypotheses that have a low probability and to predict them with the corresponding frequency. We demonstrate these features on a \lungs{} segmentation task, where each lesion has been segmented independently by four experts, and on the Cityscapes dataset, where we artificially flip labels with a certain frequency during training.

A body of work with different approaches towards probabilistic and multi-modal segmentation exists. The most common approaches provide independent pixel-wise probabilities \cite{kendall2015bayesian,kendall2017uncertainties}. 
These models induce a probability distribution by using dropout over spatial features. Whereas this strategy fulfills this line of work’s objective of quantifying the pixel-wise uncertainty, it produces inconsistent outputs.
A simple way to produce plausible hypotheses is to learn an ensemble of (deep) models \cite{lakshminarayanan2017simple}. While the outputs produced by ensembles are consistent, they are not necessarily diverse and ensembles are typically not able to learn the rare variants as their members are trained independently. In order to overcome this, several approaches train models jointly using the oracle set loss \cite{guzman2012multiple}, i.e. a loss that only accounts for the closest prediction to the ground truth. This has been explored in \cite{lee2015m} and \cite{lee2016stochastic} using an ensemble of deep networks, and in \cite{rupprecht2017learning} and \cite{Ilg2018uncertainty} using one common deep network with $M$ heads. While multi-head approaches may have the capacity to capture a diverse set of variants, they are not equipped to learn the occurrence frequencies of individual variants. Two common disadvantages of both ensembles and $M$ heads models are their ungraceful scaling to large numbers of hypotheses, and their requirement of fixing the number of allowed hypotheses at training time. Another set of approaches to produce multiple diverse solutions relies on graphical models, such as junction chains \cite{chen2013computing}, and more generally Markov Random Fields \cite{batra2012diverse,kirillov2015inferring,kirillov2015m,kirillov2016joint}. While many of the previous approaches are guaranteed to find the best diverse solutions, these are confined to structured problems whose dependencies can be described by tractable graphical models.

The task of image-to-image translation \cite{isola2017image} tackles a very similar problem: an under-constrained domain transfer of images needs to be learned. Many of the recent approaches employ generative adversarial networks (GANs) which are known to suffer from challenges such as `mode-collapse' \cite{goodfellow2016nips}.
In an attempt to solve the mode-collapse problem, the `bicycleGAN' \cite{zhu2017toward} involves a component that is similar in architecture to ours. In contrast to our proposed architecture, their model encompasses a fixed prior distribution and during training their posterior distribution is only conditioned on the output image. Very recent work on generating appearances given a shape encoding \cite{esser2018variational} also combines a \unet{} with a VAE, and was developed concurrently to ours. In contrast to our proposal, their training requires an additional pretrained VGG-net that is employed as a reconstruction loss. 
Finally, in \cite{bouchacourt2016disco} is proposed a probabilistic model for structured outputs based on optimizing the dissimilarity coefficient \cite{rao1982diversity} between the ground truth and predicted distributions. The resultant approach is assessed on the task of hand pose estimation, that is, predicting the location of 14 joints, arguably a simpler space compared to the space of segmentations we consider here. Similarly to the approach presented below, they inject latent variables at a later stage of the network architecture.

The main contributions of this work are: (1) Our framework provides consistent segmentation maps instead of pixel-wise probabilities and can therefore give a joint likelihood of modes. (2) Our model can induce arbitrarily complex output distributions including the occurrence of very rare modes, and is able to learn calibrated probabilities of segmentation modes. (3) Sampling from our model is computationally cheap. (4) In contrast to many existing applications of deep generative models that can only be qualitatively evaluated, our application and datasets allow quantitative performance evaluation including penalization of missing modes.

\section{Network Architecture and Training Procedure}

Our proposed network architecture is a combination of a conditional variational auto encoder \cite{vae1,vae2,vae3,vae4} with a \unet{} \cite{Ronneberger2015}, 
with the objective of learning
a conditional density model over segmentations, conditioned on the image.


\textbf{Sampling.} The central component of our architecture (\autoref{fig:architecture}a) is a low-dimensional latent space $\R^N$ (e.g., $N=6$, which performed best in our experiments).
Each position in this space encodes a segmentation variant.  
The `prior net', parametrized by weights $\omega$,
estimates the probability of these variants for a given input image $X$.
This prior probability distribution (called $P$ in the following) is modelled as an axis-aligned Gaussian with mean $\vec{\mu}_{\text{prior}}(X; \omega) \in \R^N$ and variance $\vec{\sigma}_{\text{prior}}(X; \omega) \in \R^N$. 
To predict a set of $m$ segmentations we apply the network $m$ times to the same input image (only a small part of the network needs to be re-evaluated in each iteration, see below). In each iteration $i \in \{1,\dots,m\}$, we draw a random sample $\vec{z}_i \in \R^N$ from $P$ 
\begin{align}
\vec{z}_i \sim P(\cdot|X) = \mathcal{N}\left(\vec{\mu}_{\text{prior}}(X; \omega),\, \text{diag}(\vec{\sigma}_{\text{prior}}(X; \omega))\right)\,,
\label{eq:sample_z}
\end{align}
broadcast the sample to an $N$-channel feature map with the same shape as the segmentation map, and concatenate this feature map to the last activation map of a \unet{} (the \unet{} is parameterized by weights $\theta$). A function $f_{\text{comb.}}$ composed of three subsequent $1\times1$ convolutions ($\psi$ being the set of their weights) combines the information and maps it to the desired number of classes. 
The output, $S_i$, is the segmentation map corresponding to point $\vec{z}_i$ in the latent space: 
\begin{align}
S_i = f_{\text{comb.}}\left(f_{\text{U-Net}}(X; \theta), \vec{z}_i; \psi\right)\,.
\label{eq:sample_y}
\end{align}
Notice that when drawing $m$ samples for the same input image, we can reuse the output of the prior net and the feature activations of the U-Net. Only the function $f_{\text{comb.}}$ needs to be re-evaluated $m$ times.

\textbf{Training.} The networks are trained with the standard training procedure for conditional VAEs (\autoref{fig:architecture}b), i.e. by minimizing the variational lower bound (\autoref{eq:elbo}). 
The main difference with respect to training a deterministic segmentation model, is that the training process additionally needs to find a useful embedding of the segmentation variants in the latent space. This is solved by introducing a `posterior net', parametrized by weights $\nu$,  that learns to recognize a segmentation variant (given the raw image $X$ and the ground truth segmentation $Y$) and to map this to a position $\vec{\mu}_{\text{post}}(X, Y; \nu) \in \R^N$ with some uncertainty $\vec{\sigma}_{\text{post}}(X, Y; \nu) \in \R^N$ in the latent space. The output is denoted as posterior distribution $Q$. A sample $\vec{z}$ from this distribution, 
\begin{equation}
\vec{z} \sim Q(\cdot|X, Y) = \mathcal{N}\left(\vec{\mu}_{\text{post}}(X, Y; \nu), \text{diag}(\vec{\sigma}_{\text{post}}(X, Y; \nu))\right),
\label{eq:posterior}
\end{equation}
combined with the activation map of the \unet{} (\autoref{eq:sample_z})
 must result in a predicted segmentation $S$ identical to the ground truth segmentation $Y$ provided in the training example. A cross-entropy loss penalizes differences between $S$ and $Y$ (the cross-entropy loss arises from treating the output $S$ as the parameterization of a pixel-wise categorical distribution $P_c$).
Additionally there is a Kullback-Leibler divergence $D_\text{KL}(Q||P) = \mathbb{E}_{z \sim Q}\left[\log\, Q - \log\, P\right]$ which penalizes differences between the posterior distribution $Q$ and the prior distribution $P$. Both losses are combined as a weighted sum with a weighting factor $\beta$, as done in \cite{higgins2016beta}:
\begin{equation}
{\cal L}(Y,X) = \mathbb{E}_{z \sim Q(\cdot|Y,X)}\big[-\log\, P_c(Y|S(X,z))\big] + \beta \cdot D_\text{KL} \big(Q(z|Y,X)||P(z|X)\big).
\label{eq:elbo}
\end{equation}
The training is done from scratch with randomly initialized weights. During training, this KL loss ``pulls'' the posterior distribution (which encodes a segmentation variant) and the prior distribution towards each other. On average (over multiple training examples) the prior distribution will be modified in a way such that it ``covers'' the space of all presented segmentation variants for a specific input image\footnote{An open source re-implementation of our approach can be found at \url{https://github.com/SimonKohl/probabilistic_unet}.}.

\section{Performance Measures and Baseline Methods}
In this section we first present the metric used to assess the performance of all approaches, and then describe each competitor approach used in the comparisons.

\subsection{Performance measures}

As it is common in the semantic segmentation literature, we employ the intersection over union (IoU) as a measure to compare a pair of segmentations. However, in the present case, we not only want to compare a deterministic prediction with a unique ground truth, but rather we are interested in comparing distributions of segmentations.
To do so, we use the \textit{generalized energy distance} 
\cite{bellemare2017cramer, salimans2018improving, szekely2013energy}, which leverages distances between observations:
\begin{equation}
    D^2_{\text{GED}}(P_\text{gt}, P_\text{out}) = 2\mathbb{E}\big[d(S,Y)\big] - \mathbb{E}\big[d(S,S^{'})\big] - \mathbb{E}\big[d(Y,Y^{'})\big],
    \label{eq:energy_distance}
\end{equation}
where $d$ is a distance measure, $Y$ and $Y^{'}$ are independent samples from the ground truth distribution $P_\text{gt}$, and similarly, $S$ and $S^{'}$ are independent samples from the predicted distribution $P_\text{out}$.
The energy distance $D_{\text{GED}}$ is a metric as long as $d$ is also a metric \cite{klebanov2005n}. In our case we choose $d(x, y)=1-\text{IoU}(x, y)$, which as proved in \cite{kosub2016note,lipkus1999proof}, is a metric.
In practice, we only have access to samples from the distributions that models induce, so we rely on statistics of \autoref{eq:energy_distance}, $\hat{D}^2_{\text{GED}}$. The details about its computation for each experiment are presented in \autoref{app:metrics}.

\subsection{Baseline methods}

\begin{figure}[htbp]
\centering
\includegraphics[width=\textwidth]{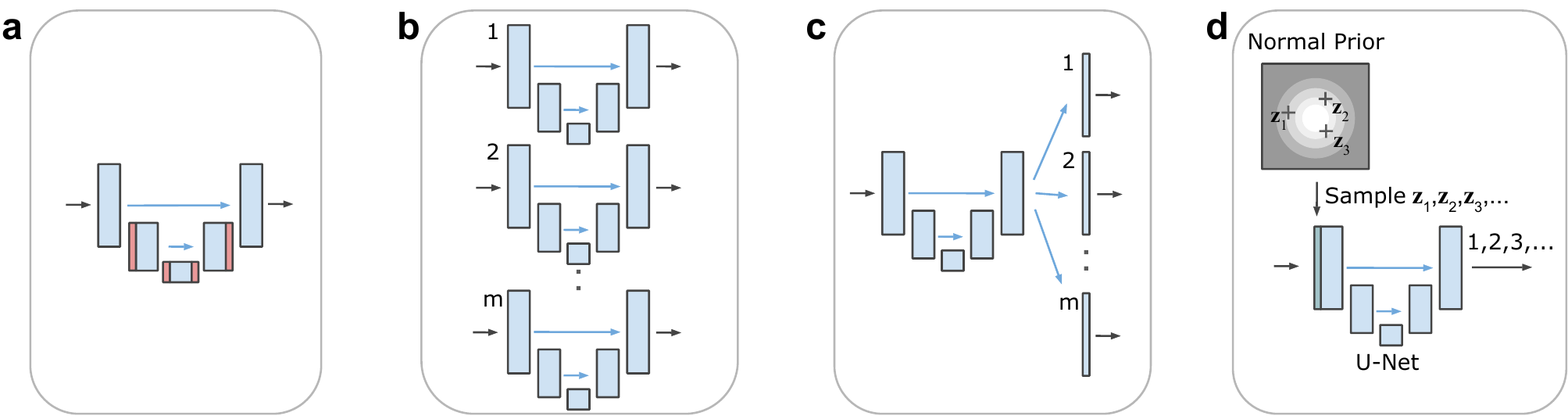}
\captionsetup{format=hang}
\caption{Baseline architectures. Arrows: flow of operations; blue blocks: feature maps; red blocks: feature maps with dropout; green block broadcasted latents. Note that the number of feature map blocks shown is reduced for clarity of presentation. (\textbf{a}) \dropoutunet. (\textbf{b}) \ensemble. (\textbf{c}) \mheads. (\textbf{d}) \itoivae.}
\label{fig:baselines}
\end{figure}

With the aim of providing context for the performance of our proposed approach we compare against a range of baselines. To the best of our knowledge there exists no other work that has considered capturing a distribution over multi-modal segmentations and has measured the agreement with such a distribution. For fair comparison, we train the baseline models whose architectures are depicted in \autoref{fig:baselines} in the exact same manner as we train ours. The baseline methods all involve the same \unet{} architecture, i.e. they share the same core component and thus employ comparable numbers of learnable parameters in the segmentation tasks.

\textbf{\dropoutunet} (\autoref{fig:baselines}a). Our `\dropoutunet' baselines follow the Bayesian segnet's \cite{kendall2015bayesian} proposition: we dropout the activations of the respective incoming layers of the three inner-most encoder and decoder blocks with a dropout probability of $p = 0.5$ during training as well as when sampling.

\textbf{\ensemble} (\autoref{fig:baselines}b). We report results for ensembles with the number of members matching the required number of samples (referred to as `\ensemble'). The original deterministic variant of the \unet{} is the 1-sample corner case of an ensemble. 

\textbf{\mheads} (\autoref{fig:baselines}c). Aiming for diverse semantic segmentation outputs, the works of \cite{rupprecht2017learning} and \cite{Ilg2018uncertainty} propose to branch off \textit{M} heads after the last layer of a deep net each of which contributes one output variant. An adjusted cross-entropy loss that adaptively assigns heads to ground-truth hypotheses is employed to promote diversity while reducing the risk of idle heads: the loss of the best performing head is weighted with a factor of $1 - \epsilon$, while the remaining heads each contribute with a weight of $\epsilon/(M-1)$ to the loss. For our `\mheads' baselines we again employ a \unet{} core and set $\epsilon = 0.05$ as proposed by \cite{rupprecht2017learning}. In order to allow for the evaluation of 4, 8 and 16 samples, we train \mheads{} models with the corresponding number of heads.

\textbf{\itoivae} (\autoref{fig:baselines}d). In \cite{zhu2017toward} the authors propose a \unet{} VAE-GAN hybrid for multi-modal image-to-image translation, that owes its stochasticity to normal distributed latents that are broadcasted and fed into the encoder path of the \unet{}. In order to deal with the complex solution space in image-to-image translation tasks, they employ an adversarial discriminator as additional supervision alongside a reconstruction loss. In the fully supervised setting of semantic segmentation such an additional learning signal is however not necessary and we therefore train with a cross-entropy loss only. In contrast to our proposition, this baseline, which we refer to as the `\itoivae', employs a prior that is not conditioned on the input image (a fixed normal distribution) and a posterior net that is not conditioned on the input either.

In all cases we examine the models' performance when drawing a different number of samples (1, 4, 8 and 16) from each of them.

\section{Results}
A quantitative evaluation of multiple segmentation predictions per image requires annotations from multiple labelers. Here we consider two datasets: The LIDC-IDRI dataset~\cite{armato2015,armato2011lung,clark2013cancer} which contains 4 annotations per input, and the Cityscapes dataset~\cite{cordts2016cityscapes}, which we artificially modify by adding \textit{synonymous classes} to introduce uncertainty in the way concepts are labelled.

\subsection{Lung abnormalities segmentation}
\label{sec:lidc_dataset}
The LIDC-IDRI dataset~\cite{armato2015,armato2011lung,clark2013cancer} contains 1018 lung CT scans from 1010 lung patients with manual lesion segmentations from four experts. This dataset is a good representation of the typical ambiguities that appear in CT scans. For each scan, 4 radiologists (from a total of 12) provided annotation masks for lesions that they independently detected and considered to be abnormal. We use the masks resulting from a second reading in which the radiologists were shown the anonymized annotations of the others and were allowed to make adjustments to their own masks.

\begin{figure}[tbp]
\centering
\includegraphics[width=1.\textwidth]{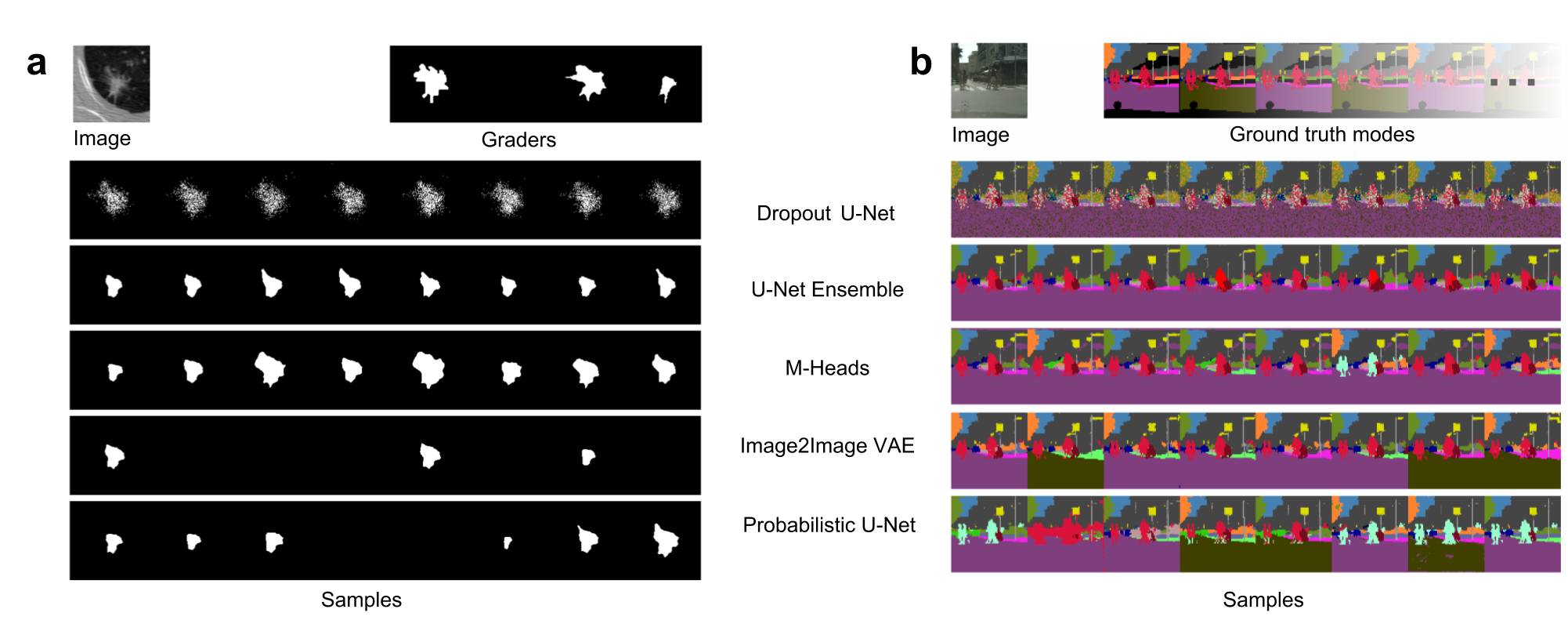}
\captionsetup{format=hang}
\caption{Qualitative results. The first row shows the input image and the ground truth segmentations. The following rows show results from the baselines and from our proposed method. (\textbf{a}) lung CT scan from the LIDC test set. Ground truth: 4 graders. (\textbf{b}) Cityscapes. Images cropped to squares for ease of presentation. Ground truth: 32 artificial modes. 
Best viewed in colour.}
\label{fig:lidc_samples}
\end{figure}

For our experiments we split this dataset into a training set composed of 722 patients, a validation set composed of 144 patients, and a test set composed of the remaining 144 patients.
We then resampled the CT scans to $\SI{0.5}{\milli\meter} \times \SI{0.5}{\milli\meter}$  in-plane resolution (the original resolution is between \SI{0.461}{\milli\meter} and \SI{0.977}{\milli\meter}, \SI{0.688}{\milli\meter} on average) and cropped 2D images ($180 \times 180$ pixels) centered at the lesion positions. The lesion positions are those where at least one of the experts segmented a lesion. 
By cropping the scans, the resultant task is in isolation not directly clinically relevant. However, this allows us to ignore the vast areas in which all labelers agree, in order to focus on those where there is uncertainty.
This resulted in 8882 images in the training set, 1996 images in the validation set and 1992 images in the test set. Because the experts can disagree whether the lesion is abnormal tissue, up to 3 masks per image can be empty. \autoref{fig:lidc_samples}a shows an example of such lesion-centered images and the masks provided by 4 graders.

\begin{figure}[tbp]

\includegraphics[width=1.\textwidth]{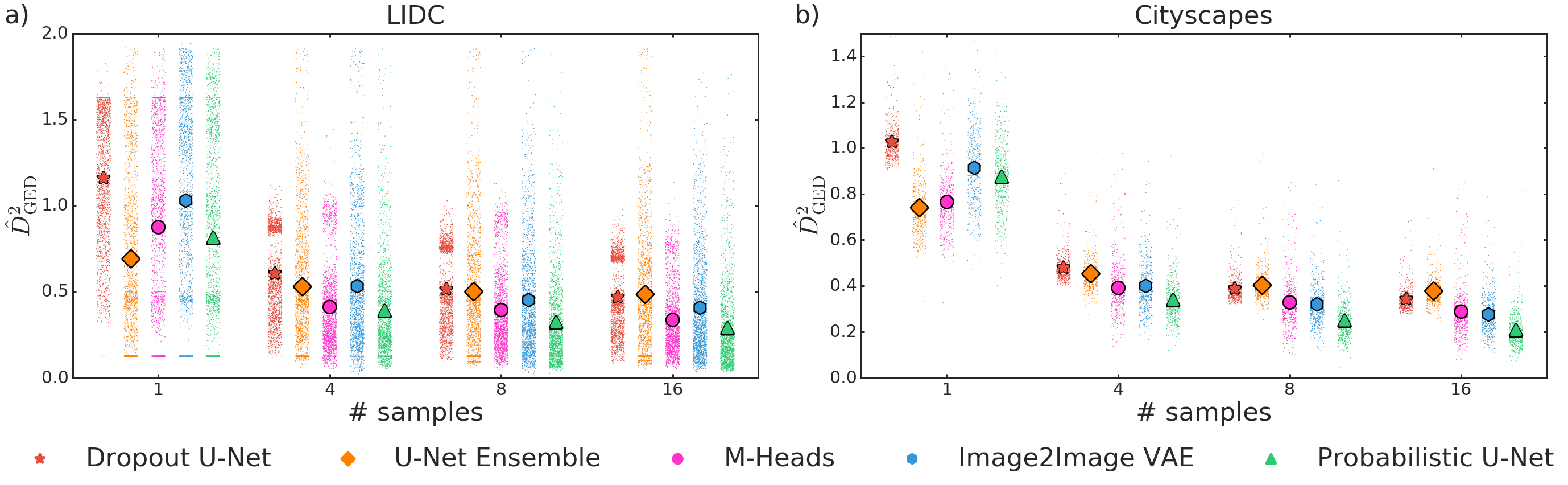}
\captionsetup{format=hang}
\caption{Comparison of approaches using the squared energy distance. Lower energy distances correspond to better agreement between predicted distributions  and ground truth distribution of segmentations. The symbols that overlay the distributions of data points mark the mean performance. (\textbf{a}) Performance on \lungs{} segmentation on our LIDC-IDRI test-set. (\textbf{b}) Performance on the official Cityscapes validation set (our test set).}
\label{fig:distance_results}
\end{figure}

As all models share the same \unet{} core component and for fairness and ease of comparability, we let all models undergo the same training schedule, which is detailed in \autoref{app:details_lidc}.

In order to grasp some intuition about the kind of samples produced by each model, we show in \autoref{fig:lidc_samples}a, as well as in \autoref{app:sampling_lidc}, representative results for the baseline methods and our proposed \probunet. 
\autoref{fig:distance_results}a shows the squared generalized energy distance $\hat{D}^{2}_{\text{GED}}$ for all models as a function of the number of samples. The data accumulations visible as horizontal stripes are owed to the existence of empty ground-truth masks. The energy distance on the 1992 images large \lungs{} test set, decreases for all models as more samples are drawn indicating an improved matching of the ground-truth distribution as well as enhanced sample diversity. Our proposed \probunet{} outperforms all baselines when sampling 4, 8 and 16 times (numerical results can be found in \autoref{table:lidc_results}). The performance at 16 samples is found significantly higher than that of the baselines $(p$-value $\sim \mathcal{O}(10^{-13}))$, according to the Wilcoxon signed-rank test. Finally, in \autoref{app:ambiguity} we show the results of an experiment regarding the capacity different models have to distinguish between  unambiguous and ambiguous instances (i.e. instances where graders disagree on the presence of a lesion).

\subsection{Cityscapes semantic segmentation} 
\label{sec:cityscapes_task}
As a second dataset we use the Cityscapes dataset \cite{cordts2016cityscapes}. It contains images of street scenes taken from a car with corresponding semantic segmentation maps. A total of 19 different semantic classes are labelled. Based on this dataset we designed a task that allows full control of the ambiguities: we create ambiguities by artificial random flips of five classes to newly introduced classes. We flip `sidewalk' to `sidewalk 2' with a probability of $8/17$, `person' to `person 2' with a probability of $7/17$, `car' to `car 2' with $6/17$, `vegetation' to `vegetation 2' with $5/17$ and `road' to `road 2' with probability $4/17$. This choice yields distinct probabilities for the ensuing $2^5 = 32$ discrete modes with probabilities ranging from 10.9\% (all unflipped) down to 0.5\% (all flipped). The official training dataset with fine-grained annotation labels comprises 2975 images and the validation dataset contains 500 images. We employ this offical validation set as a test set to report results on, and split off 274 images (corresponding to the 3 cities of Darmstadt, Mönchengladbach and Ulm) from the official training set as our internal validation set. 
As in the previous experiment, in this task we use a similar setting for the training processes of all approaches, which we present in detail in \autoref{app:details_cityscapes}.

\autoref{fig:lidc_samples}b shows samples of each approach in the comparison given one input image. In \autoref{app:sampling_cityscapes} we show further samples of other images, produced by our approach. \autoref{fig:distance_results}b shows that the \probunet{} on the Cityscapes task outperforms the baseline methods when sampling 4, 8 and 16 times in terms of the energy distance (numerical results can be found in \autoref{table:cityscapes_results}). This edge in segmentation performance at 16 samples is highly significant according to the Wilcoxon signed-rank test $(p$-value $\sim \mathcal{O}(10^{-77}))$. 
We have also conducted ablation experiments in order to explore which elements of our architecture contribute to its performance. These were (1) Fixing the prior, (2) Fixing the prior, and not using the context in the posterior and (3) Injecting the latent features at the beginning of the U-Net. Each of these variations resulted in a lower performance. Detailed results can be found in \autoref{app:ablation}.

\paragraph{Reproducing the segmentation probabilities.}
\label{sec:cityscapes_quantitative}

In the Cityscapes segmentation task, we can provide further analysis by leveraging our knowledge of the underlying conditional distribution that we have set by design.
In particular we compare the frequency with which every model predicts each mode, to the corresponding ground truth probability of that mode. 
To compute the frequency of each mode by each model, we draw 16 samples from that model for all images in the test set. Then we count the number of those samples that have that mode as the closest
(using 1-IoU as the distance function).

\begin{figure}[bp]
\centering
\includegraphics[width=1.\textwidth]{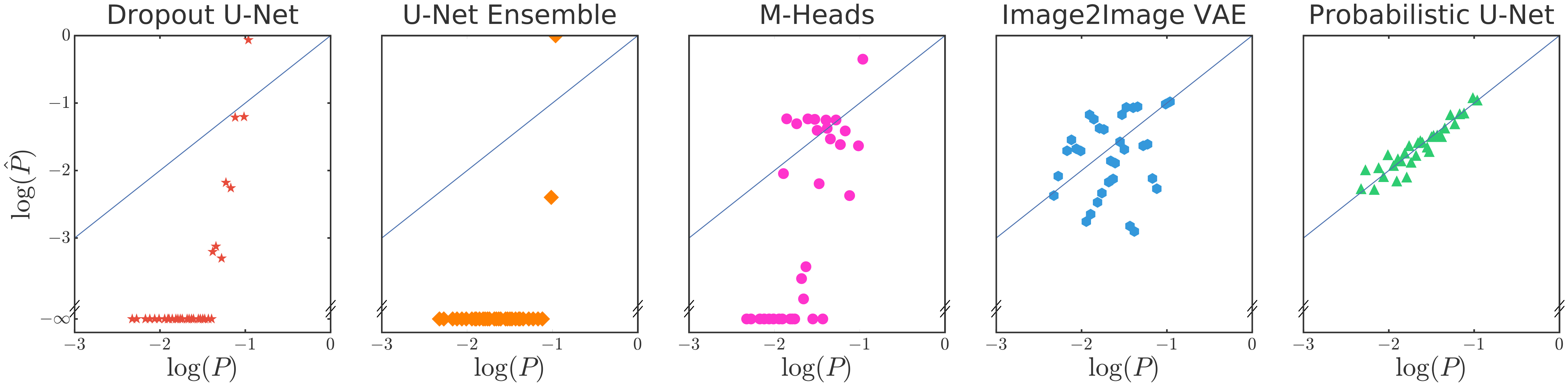}
\captionsetup{format=hang}
\caption{Reproduction of the probabilities of the segmentation modes on the Cityscapes task. The artificial flipping of 5 classes results in 32 modes with different ground truth probability (x-axis). The y-axis shows the frequency of how often the model predicted this variant in the whole test set. Agreement with the bisector line indicates calibration quality.}
\label{fig:log_frequencies}
\end{figure}

In \autoref{fig:log_frequencies} (and Figs.~\ref{fig:freq_ensemble_and_dropout}, \ref{fig:freq_m_heads_and_image2image}, \ref{fig:freq_prob_unet} in \autoref{app:freqs}) we report the mode-wise frequencies for all 32 modes in the Cityscape task and show that the \probunet{} is the only model in this comparison that is able to closely capture the frequencies of a large combinatorial space of hypotheses including very rare modes, thus supplying calibrated likelihoods of modes. The \itoivae{} is the only model among competitors that picks up on all variants, but the frequencies are far off as can be seen in its deviation from the bisector line in blue. The other baselines perform worse still in that all of them fail to represent modes and the modes they do capture do not match the expected frequencies.

\subsection{Analysis of the Latent Space}
The embedding of the segmentation variants in a low-dimensional latent space allows a qualitative analysis of the internal representation of our model. For a 2D or 3D latent space we can directly visualize where the segmentation variants get assigned. See \autoref{app:latent_space} for details.

\section{Discussion and conclusions}

Our first set of experiments demonstrates that our proposed architecture provides consistent segmentation maps that closely match the multi-modal ground-truth distributions given by the expert graders in the \lungs{} task and by the combinatorial ground-truth segmentation modes in the Cityscapes task. The employed IoU-based energy distance measures whether the models' individual samples are both coherent as well as whether they are produced with the expected frequencies. It not only penalizes predicted segmentation variants that are far away from the ground truth, but also penalizes missing variants. On this task the \probunet{} is able to significantly outperform the considered baselines, indicating its capability to model the joint likelihood of segmentation variants. 

The second type of experiments demonstrates that our model scales to  complex output distributions including the occurrence of very rare modes. With 32 discrete modes of largely differing occurrence likelihoods (0.5\% to 10.9\%), the Cityscapes task requires the ability to closely match complex data distributions. Here too our model performs best and picks the segmentation modes very close to the expected frequencies, all the way into the regime of very unlikely modes, thus defying mode-collapse and exhibiting excellent probability calibration. As an additional advantage our model scales to such large numbers of modes without requiring any prior assumptions on the number of modes or hypotheses.

The lower performance of the baseline models relative to our proposition can be attributed to design choices of these models. While the \textit{\dropoutunet{}} successfully models the pixel-wise data distribution (\autoref{fig:freq_ensemble_and_dropout}a bottom right, in the Appendix), such pixel-wise mixtures of variants can not be valid hypotheses in themselves (see \autoref{fig:lidc_samples}). The \textit{\ensemble}'s members are trained independently and each of them can only learn the most likely segmentation variant as attested to by \autoref{fig:freq_ensemble_and_dropout}b. In contrast to that the closely related \textit{\mheads{}} model can pick up on multiple discrete segmentation modes, due to the joint training procedure that enables diversity. The training does however not allow to correctly represent frequencies and requires knowledge of the number of present variants (see \autoref{fig:freq_m_heads_and_image2image}a, in the Appendix). Furthermore neither the \ensemble, nor the \mheads{} can deal with the combinatorial explosion of segmentation variants when multiple aspects vary independently of each other. 
The \textit{\itoivae{}} shares similarities with our model, but as its prior is fixed and not conditioned on the input image, it can not learn to capture variant frequencies by allocating corresponding probability mass to the respective latent space regions. \autoref{fig:lidc_image2image_samples} in the Appendix shows a severe miss-calibration of variant likelihoods on the \lungs{} task that is also reflected in its corresponding energy distance. Furthermore, in this architecture, the latent samples are fed into the \unet's encoder path, while we feed in the samples just after the decoder path. This design choice in the \itoivae{} requires the model to carry the latent information all the way through the \unet{} core, while simultaneously performing the recognition required for segmentation, which might additionally complicate training (see analysis in \autoref{app:ablation}). Beside that, our design choice of late injection has the additional advantage that we can produce a large set of samples for a given image at a very low computational cost: for each new sample from the latent space only the network part after the injection needs to be re-executed to produce the corresponding segmentation map (this bears similarity to the approach taken in \cite{bouchacourt2016disco}, where a generative model is employed to model hand pose estimation).

Aside from the ability to capture arbitrary modes with their corresponding probability conditioned on the input, our proposed \textit{\probunet{}} allows to inspect its latent space. This is because as opposed to e.g. GAN-based approaches, VAE-like models explicitly parametrize distributions, a characteristic that grants direct access to the corresponding likelihood landscape. \autoref{app:latent_space} discusses how the \probunet{} chooses to structure its latent spaces.


Compared to aforementioned concurrent work for image-to-image tasks \cite{esser2018variational}, our model disentangles the prior and the segmentation net. This can be of particular relevance in medical imaging, where processing 3D scans is common. In this case it is desirable to condition on the entire scan, while retaining the possibility to process the scan tile by tile in order to be able to process large volumes with  large models with a limited amount of GPU memory.  

On a more general note, we would like to remark that current image-to-image translation tasks only allow subjective (and expensive) performance evaluations, as it is typically intractable to assess the entire solution space. For this reason surrogate metrics such as the inception score based on the evaluation via a separately trained deep net are employed \cite{salimans2016improved}. The task of multi-modal semantic segmentation, which we consider here, allows for a direct and thus perhaps more meaningful manner of performance evaluation and could help guide the design of future generative architectures. 

All in all we see a large field where our proposed \probunet{} can replace the currently applied deterministic \unet s. Especially in the medical domain, with its often ambiguous images and highly critical decisions that depend on the correct interpretation of the image, our model's segmentation hypotheses and their likelihoods could 1) inform diagnosis/classification probabilities or 2) guide steps to resolve ambiguities. Our method could prove useful beyond explicitly multi-modal tasks, as the inspectability of the \probunet's latent space could yield insights for many segmentation tasks that are currently treated as a uni-modal problem. 

\section{Acknowledgements}
The authors would like to thank Mustafa Suleyman, Trevor Back and the whole DeepMind team for their exceptional support, and Shakir Mohamed and Andrew Zisserman for very helpful comments and discussions. The authors acknowledge the National Cancer Institute and the Foundation for the National Institutes of Health, and their critical role in the creation of the free publicly available LIDC/IDRI Database used in this study. 

{
\small

\bibliographystyle{splncs}
\bibliography{egbib}

\begin{thebibliography}{10}

\bibitem{lee2016stochastic}
Lee, S., Prakash, S.P.S., Cogswell, M., Ranjan, V., Crandall, D., Batra, D.:
\newblock Stochastic multiple choice learning for training diverse deep
  ensembles.
\newblock In: Advances in Neural Information Processing Systems. (2016)
  2119--2127

\bibitem{vae1}
Kingma, D.P., Welling, M.:
\newblock Auto-encoding variational bayes.
\newblock In: Proceedings of the 2nd international conference on Learning
  Representations (ICLR). (2013)

\bibitem{vae2}
Jimenez~Rezende, D., Mohamed, S., Wierstra, D.:
\newblock Stochastic backpropagation and approximate inference in deep
  generative models.
\newblock In: Proceedings of the 31st International Conference on Machine
  Learning (ICML). (2014)

\bibitem{vae3}
Kingma, D.P., Jimenez~Rezende, D., Mohamed, S., Welling, M.:
\newblock Semi-supervised learning with deep generative models.
\newblock In: Neural Information Processing Systems (NIPS). (2014)

\bibitem{vae4}
Sohn, K., Lee, H., Yan, X.:
\newblock Learning structured output representation using deep conditional
  generative models.
\newblock In: Advances in Neural Information Processing Systems. (2015)
  3483--3491

\bibitem{Ronneberger2015}
Ronneberger, O., Fischer, P., Brox, T.:
\newblock U-net: Convolutional networks for biomedical image segmentation.
\newblock In: Medical Image Computing and Computer-Assisted Intervention
  (MICCAI) 2015. Volume 9351 of LNCS., Springer (2015)  234--241

\bibitem{kendall2015bayesian}
Kendall, A., Badrinarayanan, V., Cipolla, R.:
\newblock Bayesian segnet: Model uncertainty in deep convolutional
  encoder-decoder architectures for scene understanding.
\newblock arXiv preprint arXiv:1511.02680 (2015)

\bibitem{kendall2017uncertainties}
Kendall, A., Gal, Y.:
\newblock What uncertainties do we need in bayesian deep learning for computer
  vision?
\newblock In: Advances in Neural Information Processing Systems. (2017)
  5580--5590

\bibitem{lakshminarayanan2017simple}
Lakshminarayanan, B., Pritzel, A., Blundell, C.:
\newblock Simple and scalable predictive uncertainty estimation using deep
  ensembles.
\newblock In: Advances in Neural Information Processing Systems. (2017)
  6405--6416

\bibitem{guzman2012multiple}
Guzman-Rivera, A., Batra, D., Kohli, P.:
\newblock Multiple choice learning: Learning to produce multiple structured
  outputs.
\newblock In: Advances in Neural Information Processing Systems. (2012)
  1799--1807

\bibitem{lee2015m}
Lee, S., Purushwalkam, S., Cogswell, M., Crandall, D., Batra, D.:
\newblock Why m heads are better than one: Training a diverse ensemble of deep
  networks.
\newblock arXiv preprint arXiv:1511.06314 (2015)

\bibitem{rupprecht2017learning}
Rupprecht, C., Laina, I., DiPietro, R., Baust, M., Tombari, F., Navab, N.,
  Hager, G.D.:
\newblock Learning in an uncertain world: Representing ambiguity through
  multiple hypotheses.
\newblock In: International Conference on Computer Vision (ICCV). (2017)

\bibitem{Ilg2018uncertainty}
Ilg, E., {\c{C}}i{\c{c}}ek, {\"O}., Galesso, S., Klein, A., Makansi, O.,
  Hutter, F., Brox, T.:
\newblock Uncertainty estimates for optical flow with multi-hypotheses
  networks.
\newblock arXiv preprint arXiv:1802.07095 (2018)

\bibitem{chen2013computing}
Chen, C., Kolmogorov, V., Zhu, Y., Metaxas, D., Lampert, C.:
\newblock Computing the m most probable modes of a graphical model.
\newblock In: Artificial Intelligence and Statistics. (2013)  161--169

\bibitem{batra2012diverse}
Batra, D., Yadollahpour, P., Guzman-Rivera, A., Shakhnarovich, G.:
\newblock Diverse m-best solutions in markov random fields.
\newblock In: European Conference on Computer Vision, Springer (2012)  1--16

\bibitem{kirillov2015inferring}
Kirillov, A., Savchynskyy, B., Schlesinger, D., Vetrov, D., Rother, C.:
\newblock Inferring m-best diverse labelings in a single one.
\newblock In: Proceedings of the IEEE International Conference on Computer
  Vision. (2015)  1814--1822

\bibitem{kirillov2015m}
Kirillov, A., Shlezinger, D., Vetrov, D.P., Rother, C., Savchynskyy, B.:
\newblock M-best-diverse labelings for submodular energies and beyond.
\newblock In: Advances in Neural Information Processing Systems. (2015)
  613--621

\bibitem{kirillov2016joint}
Kirillov, A., Shekhovtsov, A., Rother, C., Savchynskyy, B.:
\newblock Joint m-best-diverse labelings as a parametric submodular
  minimization.
\newblock In: Advances in Neural Information Processing Systems. (2016)
  334--342

\bibitem{isola2017image}
Isola, P., Zhu, J.Y., Zhou, T., Efros, A.A.:
\newblock Image-to-image translation with conditional adversarial networks.
\newblock arXiv preprint (2017)

\bibitem{goodfellow2016nips}
Goodfellow, I.:
\newblock Nips 2016 tutorial: Generative adversarial networks.
\newblock arXiv preprint arXiv:1701.00160 (2016)

\bibitem{zhu2017toward}
Zhu, J.Y., Zhang, R., Pathak, D., Darrell, T., Efros, A.A., Wang, O.,
  Shechtman, E.:
\newblock Toward multimodal image-to-image translation.
\newblock In: Advances in Neural Information Processing Systems. (2017)
  465--476

\bibitem{esser2018variational}
Esser, P., Sutter, E., Ommer, B.:
\newblock A variational u-net for conditional appearance and shape generation.
\newblock arXiv preprint arXiv:1804.04694 (2018)

\bibitem{bouchacourt2016disco}
Bouchacourt, D., Mudigonda, P.K., Nowozin, S.:
\newblock Disco nets: Dissimilarity coefficients networks.
\newblock In: Advances in Neural Information Processing Systems. (2016)
  352--360

\bibitem{rao1982diversity}
Rao, C.R.:
\newblock Diversity and dissimilarity coefficients: a unified approach.
\newblock Theoretical population biology \textbf{21}(1) (1982)  24--43

\bibitem{higgins2016beta}
Higgins, I., Matthey, L., Pal, A., Burgess, C., Glorot, X., Botvinick, M.,
  Mohamed, S., Lerchner, A.:
\newblock beta-vae: Learning basic visual concepts with a constrained
  variational framework.
\newblock In: International Conference on Learning Representations. (2017)

\bibitem{bellemare2017cramer}
Bellemare, M.G., Danihelka, I., Dabney, W., Mohamed, S., Lakshminarayanan, B.,
  Hoyer, S., Munos, R.:
\newblock The cramer distance as a solution to biased wasserstein gradients.
\newblock arXiv preprint arXiv:1705.10743 (2017)

\bibitem{salimans2018improving}
Salimans, T., Zhang, H., Radford, A., Metaxas, D.:
\newblock Improving gans using optimal transport.
\newblock arXiv preprint arXiv:1803.05573 (2018)

\bibitem{szekely2013energy}
Sz{\'e}kely, G.J., Rizzo, M.L.:
\newblock Energy statistics: A class of statistics based on distances.
\newblock Journal of statistical planning and inference \textbf{143}(8) (2013)
  1249--1272

\bibitem{klebanov2005n}
Klebanov, L.B., Bene{\v{s}}, V., Saxl, I.:
\newblock N-distances and their applications.
\newblock Charles University in Prague, the Karolinum Press (2005)

\bibitem{kosub2016note}
Kosub, S.:
\newblock A note on the triangle inequality for the jaccard distance.
\newblock arXiv preprint arXiv:1612.02696 (2016)

\bibitem{lipkus1999proof}
Lipkus, A.H.:
\newblock A proof of the triangle inequality for the tanimoto distance.
\newblock Journal of Mathematical Chemistry \textbf{26}(1-3) (1999)  263--265

\bibitem{armato2015}
Armato, I., Samuel, G., McLennan, G., Bidaut, L., McNitt-Gray, M.F., Meyer,
  C.R., Reeves, A.P., Clarke, L.P.:
\newblock Data from lidc-idri. the cancer imaging archive.
\newblock \url{http://doi.org/10.7937/K9/TCIA.2015.LO9QL9SX} (2015)

\bibitem{armato2011lung}
Armato, S.G., McLennan, G., Bidaut, L., McNitt-Gray, M.F., Meyer, C.R., Reeves,
  A.P., Zhao, B., Aberle, D.R., Henschke, C.I., Hoffman, E.A.,  et~al.:
\newblock The lung image database consortium (lidc) and image database resource
  initiative (idri): a completed reference database of lung nodules on ct
  scans.
\newblock Medical physics \textbf{38}(2) (2011)  915--931

\bibitem{clark2013cancer}
Clark, K., Vendt, B., Smith, K., Freymann, J., Kirby, J., Koppel, P., Moore,
  S., Phillips, S., Maffitt, D., Pringle, M.,  et~al.:
\newblock The cancer imaging archive (tcia): maintaining and operating a public
  information repository.
\newblock Journal of digital imaging \textbf{26}(6) (2013)  1045--1057

\bibitem{cordts2016cityscapes}
Cordts, M., Omran, M., Ramos, S., Rehfeld, T., Enzweiler, M., Benenson, R.,
  Franke, U., Roth, S., Schiele, B.:
\newblock The cityscapes dataset for semantic urban scene understanding.
\newblock In: Proceedings of the IEEE conference on computer vision and pattern
  recognition. (2016)  3213--3223

\bibitem{salimans2016improved}
Salimans, T., Goodfellow, I., Zaremba, W., Cheung, V., Radford, A., Chen, X.:
\newblock Improved techniques for training gans.
\newblock In: Advances in Neural Information Processing Systems. (2016)
  2234--2242

\bibitem{kingma2014adam}
Kingma, D.P., Ba, J.:
\newblock Adam: A method for stochastic optimization.
\newblock arXiv preprint arXiv:1412.6980 (2014)

\end{thebibliography}
}
\newpage
\begin{appendices}

\section{Visualization of latent spaces}
\label{app:latent_space}

The segmentation variants from the proposed \probunet{} correspond to latent space samples from the learned prior distribution. \autoref{fig:lidc_latent_space} and \autoref{fig:cityscapes_latent_space} below show samples from the \probunet{} for an LIDC-IDRI and a Cityscapes example respectively. The samples are arranged so as to represent their corresponding position in a 2D-plane of the respective latent space. This allows to interpret how the model ends up structuring the space to solve the given tasks. 

\subsection{Lung Abnormalities Segmentation}

\begin{figure}[htbp]
\centering
\makebox[\textwidth][c]{\includegraphics[width=1.1\textwidth]{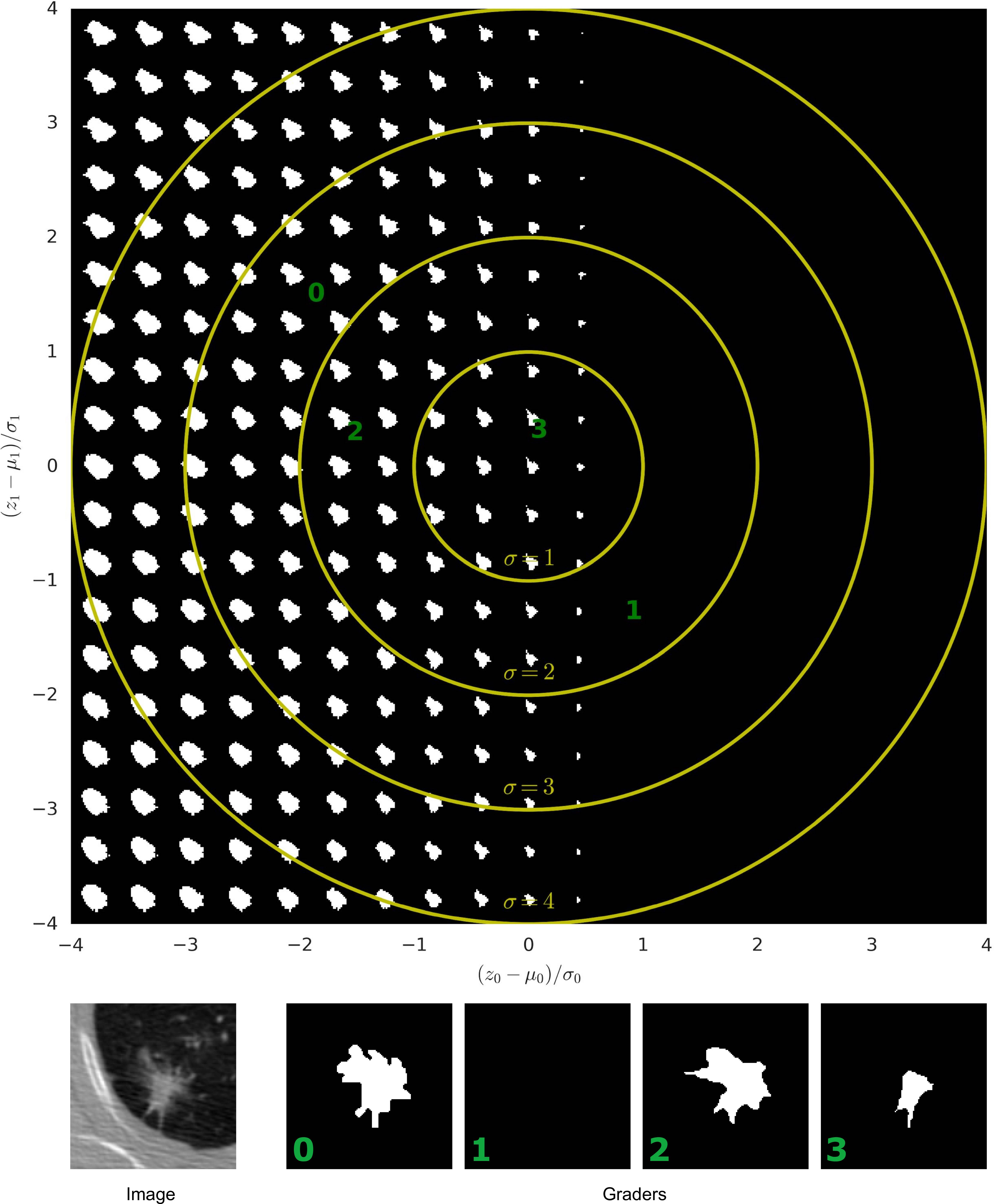}}
\captionsetup{format=hang}
\caption{Visualization of the latent space for the lung abnormalities segmentation. $19\times19$ samples for a LIDC-IDRI test set example mapped to their prior latent-space position, using our model trained with a latent space of only 2 dimensions. For ease of presentation, the latent space is re-scaled so that the prior likelihood is a spherical unit-Gaussian. The isoprobable yellow circles denote deviations from the mean in sigma. The ground-truth grader masks' posterior position in this latent space is indicated by green numbers. The input image is shown in the lower left, to the right of it, the 4 grader masks are shown.}
\label{fig:lidc_latent_space}
\end{figure}

In the LIDC-IDRI case the $z_0$-component of the prior happens to roughly encode lesion size including a transition to complete lesion absence. The probability mass allocated to absence is relatively small in the particular example, which arguably is in tune with the fact that 1 of the 4 graders assessed the image as lesion free. The $z_1$-component on the other hand appears to encode shape variations. In the training, the posterior and the prior distribution are tied by means of the KL-divergence. As a consequence they `live' in the same space and the graders (alongside the image to condition on) can be projected into the same latent space. \autoref{fig:lidc_latent_space} shows the grader's position in the form of green dots. The three graders that agree on presence, map into the 1-sigma interval of the prior, while the grader predicting absence falls just short of the 4-sigma isoprobability contour in the latent-space area that encodes absence. \autoref{fig:lidc_samples} gives more LIDC-IDRI examples with their corresponding grader masks and 16 random samples of the \probunet{}. It appears that our model agrees very well with cases for which there is inter-grader disagreement on lesion presence. For cases where the graders agree on presence, our model at times apparently shows an under-conservative prior, in the sense that uncertainty on presence can be elevated. The shape variations however are covered to a very good degree as attested by quantitative experiments above.

\subsection{Street Scene Segmentation}

\begin{figure}[htbp]
\centering
\vbox{\vspace{-20mm}
\makebox[\textwidth][c]{\includegraphics[width=1.\textwidth]{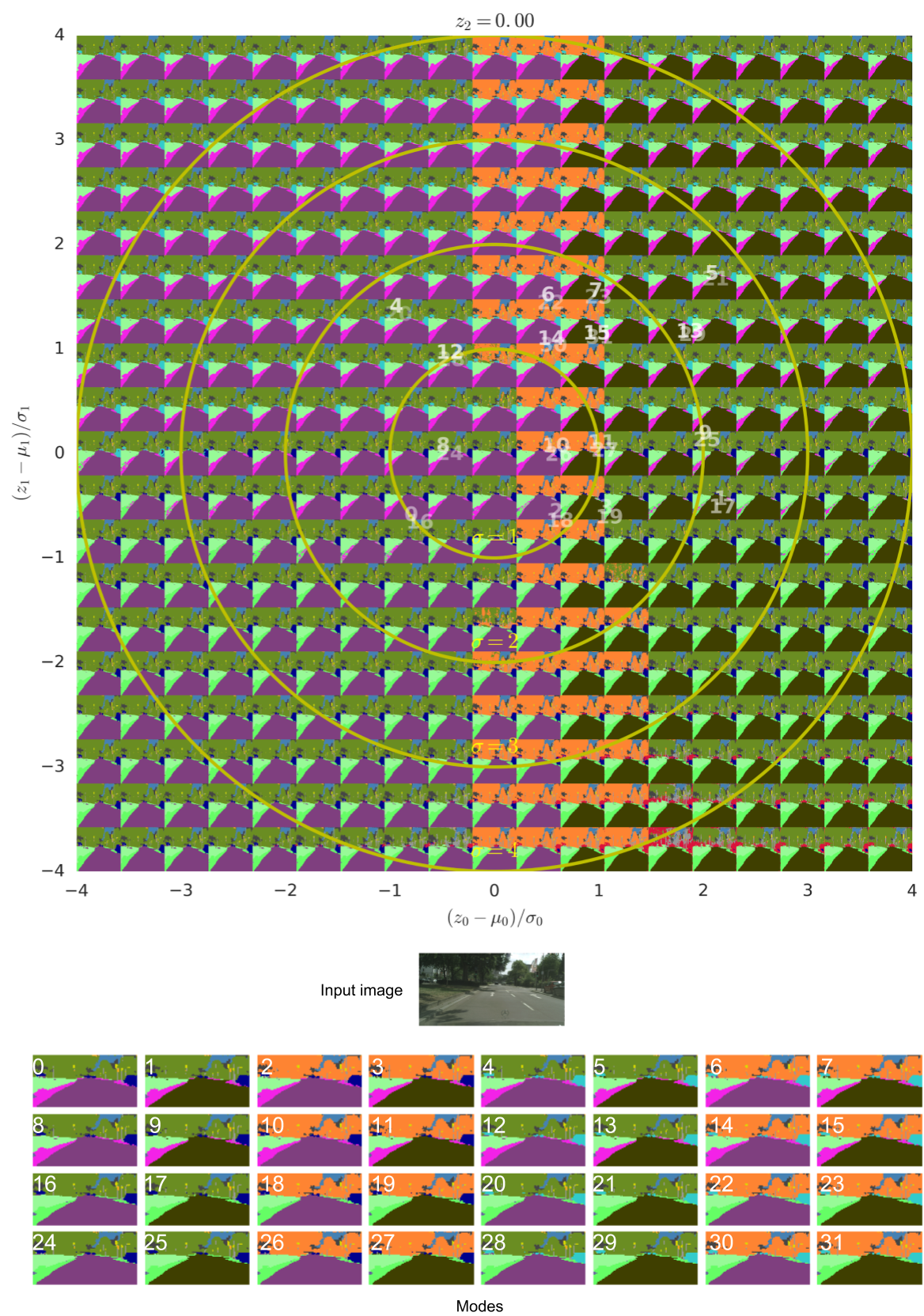}}
}
\captionsetup{format=hang}
\caption{Visualization of the latent space for the Cityscapes task. $19\times19$ samples of a Cityscapes validation set example, mapped here to their latent-space position in the $z_0$-$z_1$ plane ($z_2=0$) of the learned prior, using our model trained with a latent space of only 3 dimensions. For ease of presentation, the samples are squeezed to rectangles and the latent space is re-scaled so that the prior likelihood is a spherical unit-Gaussian. The isoprobable yellow circles denote deviations from the mean in sigma. The ground-truth grader masks' posterior position in this latent space is indicated by white numbers. (color-map as in \autoref{fig:cityscapes_samples}).}
\label{fig:cityscapes_latent_space}
\end{figure}

In the Cityscapes task we employ a latent space with more dimensions than on the lung abnormalities task in order to equip the prior with sufficient capacity to encode the grader modes. The best performing model used a 6D latent space, however, for ease of presentation the following discusses the latent structure of a 3D latent space version. \autoref{fig:cityscapes_latent_space} shows a $z_0$-$z_1$ plane of the latent space in which we again map corresponding segmentation samples, this time for a Cityscapes example. The precisely defined grader modes in the Cityscapes task can be identified with coherent regions in the latent space. As the space is 3D, not all 32 modes are fully manifest in the shown $z_2$-slice. The location of the modes is shown via white mode numbers and the degree of transparency indicated the proximity in $z_2$ relative to the shown slice. As this particular task involves discrete modes, the semantically different regions are coherent and well confined as hoped for. There however inevitably are transitions between those latent space regions that will translate to mixtures of the grader modes that cross over. Ideally these transitions are as sharp as possible relative to the order of magnitude of the prior variance, which is arguably the case. \autoref{fig:cityscapes_samples} shows Cityscapes examples with their corresponding grader masks and 16 random samples of the \probunet{}. The shown samples exhibit largely coherent variants alongside occasional variant mixtures that correspond to semantic cross overs in the latent space. As alluded to quantitatively before, the samples also appear to respect the grader variant frequencies, which are captured by structuring the latent-space under the prior in such fashion that the correct probability mass is allocated to the respective mode. In the upper boundary region of \autoref{fig:cityscapes_latent_space} improper samples are found that show miss-segmentations (although those are unlikely under the prior). The erroneously encoded modes found here are presumably attributable to the presence of inherent ambiguities in the dataset.

\clearpage

\section{Metrics}
\label{app:metrics}

In the LIDC dataset, given that we have $m=4$ ground truth samples and $n$ samples from the models, we employ the following statistic:
\begin{equation}
    \hat{D}^2_{\text{GED}}(P_\text{gt}, P_\text{out}) = 
    \frac{2}{nm}
    \sum_{i=1}^n \sum_{j=1}^m 
    d(S_i,Y_j) 
    - \frac{1}{n^2} 
    \sum_{i=1}^n \sum_{j=1}^n 
    d(S_i,S_j^{'}) 
    - \frac{1}{m^2} 
    \sum_{i=1}^m \sum_{j=1}^m
    d(Y_i,Y_j^{'}).
\end{equation}
Here $d(x, y)=1-\text{IoU}(x, y)$, where $x$ and $y$ are the predicted and ground truth masks of the lesion. In the case that both are empty masks, we define its distance to be $0$, so that the metric rewards the agreement on lesion absence. 

On the Cityscapes task, given that we have defined the settings, we have full knowledge about the ground truth distribution, which is a mixture of $M=32$ Dirac delta distributions. Hence, we do not need to sample from it, but use it directly in the estimator:
\begin{equation}
    \hat{D}^2_{\text{GED}}(P_\text{gt}, P_\text{out}) = \frac{2}{n}
    \sum_{i=1}^n \sum_{j=1}^M 
    d(S_i,Y_j)\omega_j 
    - \frac{1}{n^2} 
    \sum_{i=1}^n \sum_{j=1}^n 
    d(S_i,S_j^{'}) -
    \sum_{i=1}^M \sum_{j=1}^M
    d(Y_i,Y_j^{'})\omega_i\omega_j,
\end{equation}
where $\omega_j$ is the weight for the $j$-th mixture, which is a delta distribution containing all the density in $Y_j$. 
Here the distance $d$ depends on the average IoU of the $10$ switchable classes only. Predicting one of such classes that is not present in the ground truth leads to a $0$ score, which will be one of the terms over which we average. The computed average does not account for classes that are not present in both prediction and ground truth.


\section{How models fit the ground truth distribution}
\label{app:freqs}
\begin{figure}[htbp]
\centering
\raisebox{.70\textwidth}{a}
\includegraphics[width=.75\textwidth]{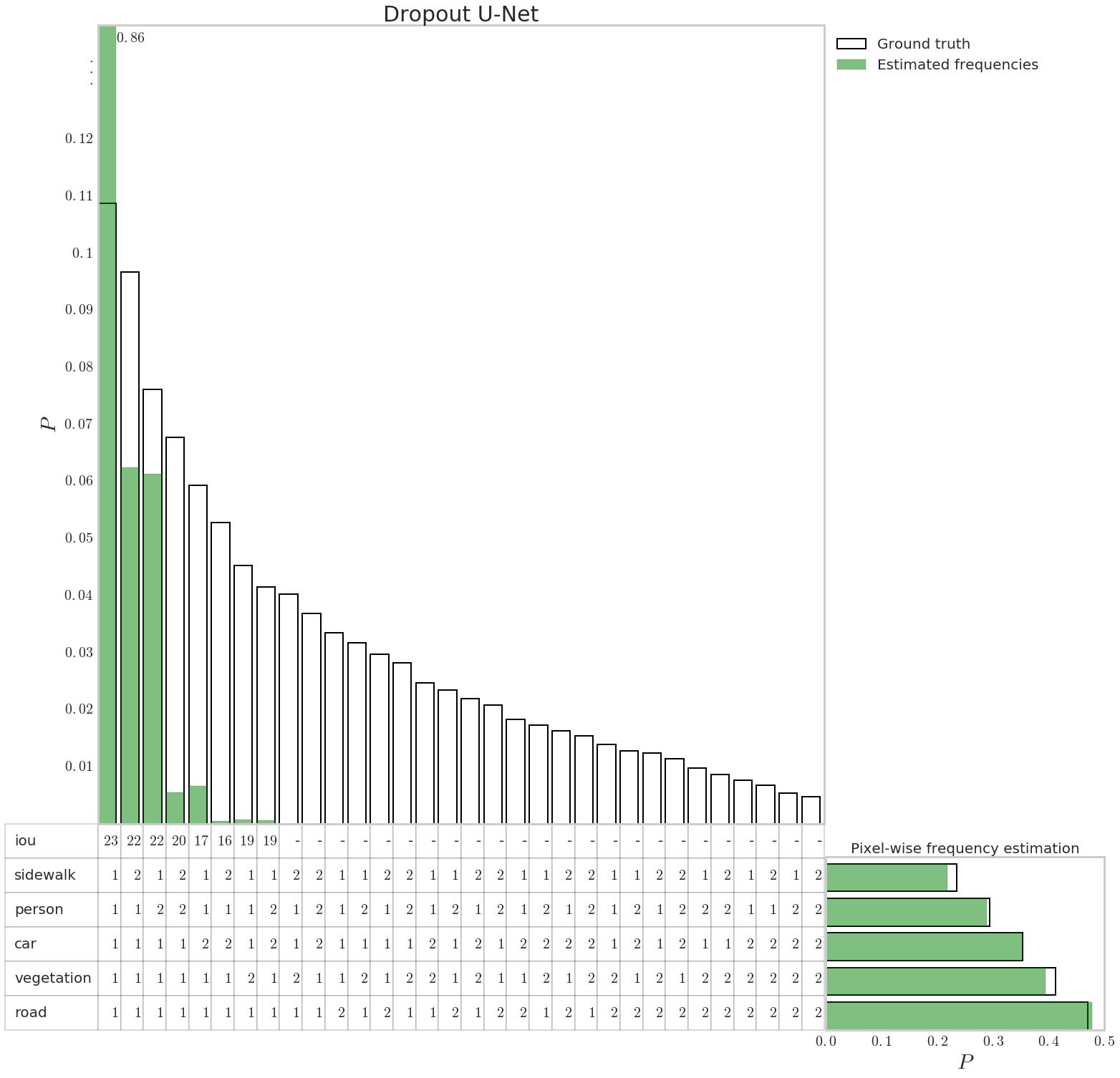}\\
\raisebox{.70\textwidth}{b}
\includegraphics[width=.75\textwidth]{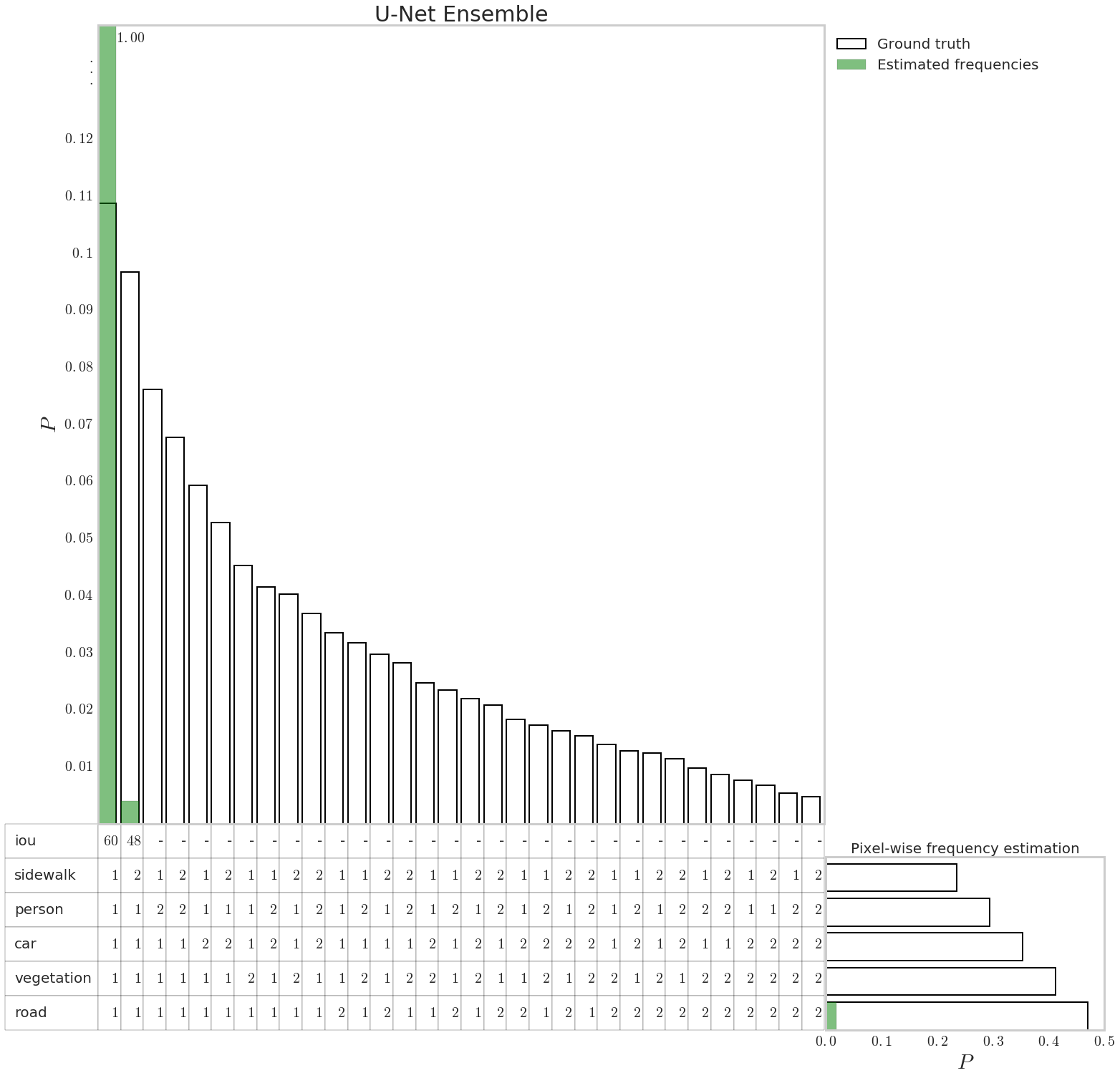}
\captionsetup{format=hang}
\caption{Reproduction of probabilities by the baselines \dropoutunet{} and \ensemble{}. The vertical histogram shows the mode-wise occurrence frequencies of samples in comparison to the ground-truth probability of the modes, and the horizontal histogram reports the pixel-wise marginal frequencies, i.e. the sampled pixel-fractions for each new stochastic class (e.g. sidewalk 2) with respect to the corresponding existing one (sidewalk).}
\label{fig:freq_ensemble_and_dropout}
\end{figure}

\begin{figure}[htbp]
\centering
\raisebox{.70\textwidth}{a}
\includegraphics[width=.75\textwidth]{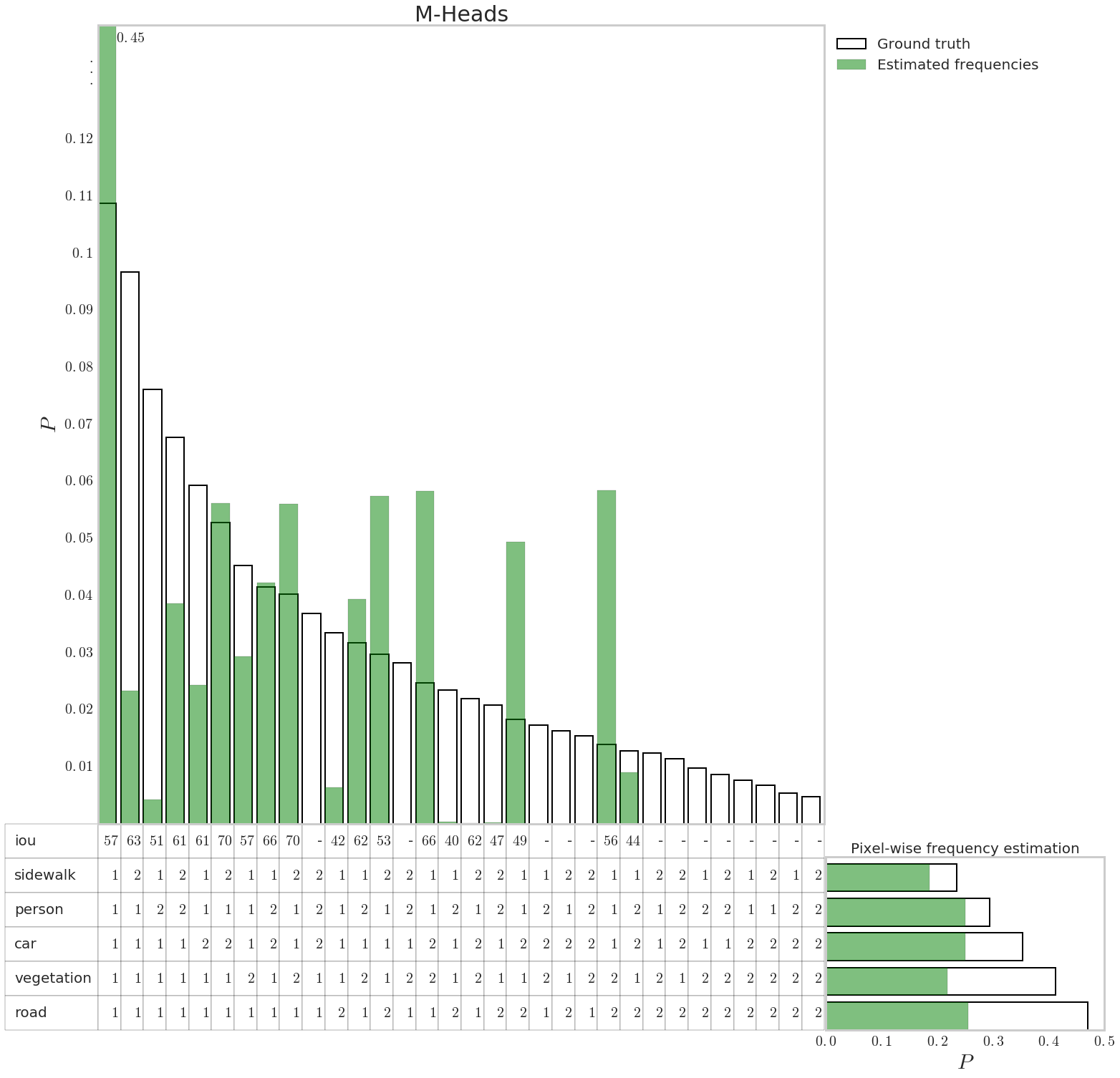}\\
\raisebox{.70\textwidth}{b}
\includegraphics[width=.75\textwidth]{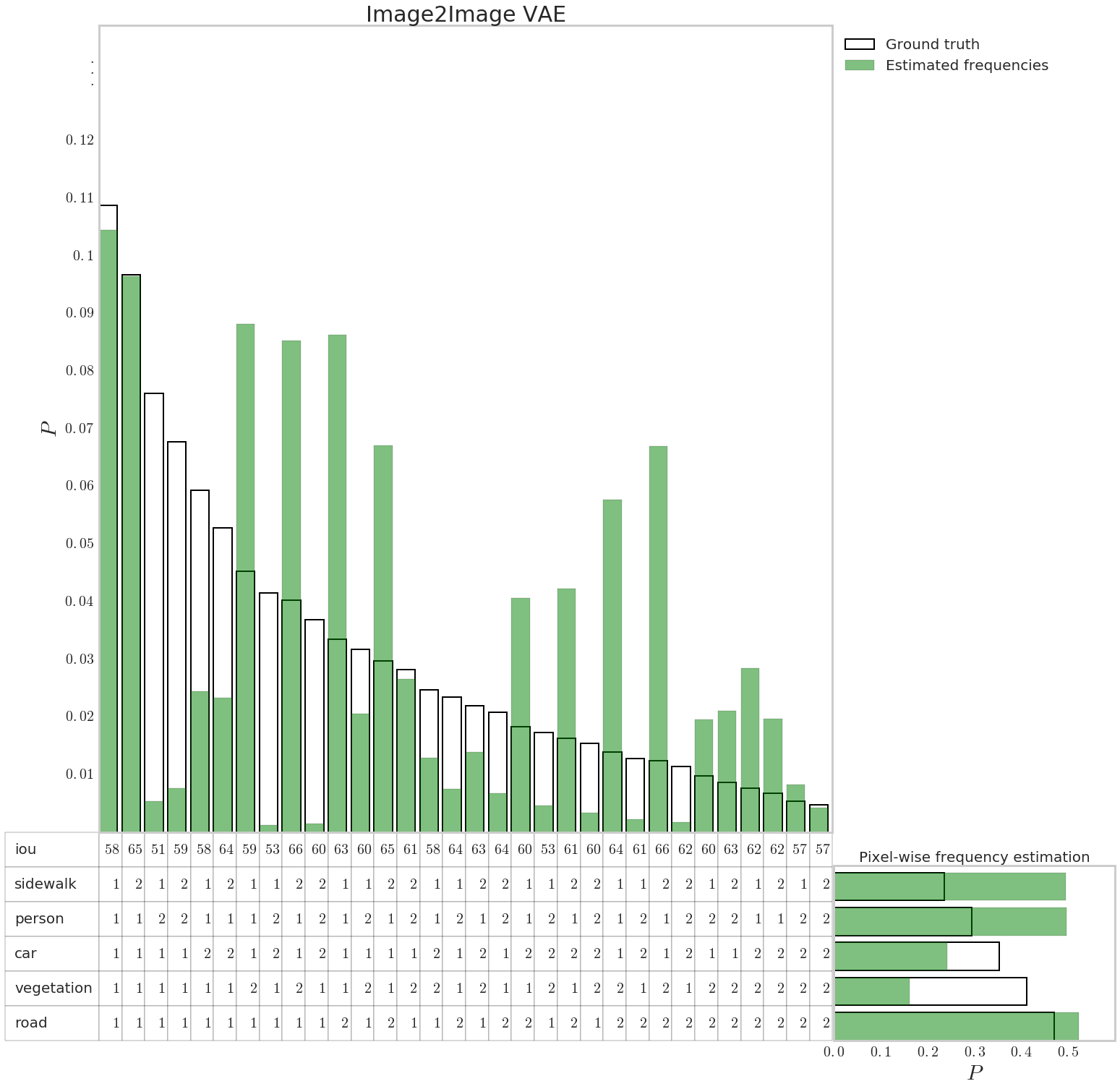}
\captionsetup{format=hang}
\caption{Reproduction of probabilities by the baselines \mheads{} and \itoivae{}. The vertical histogram shows the mode-wise occurrence frequencies of samples in comparison to the ground-truth probability of the modes, and the horizontal histogram reports the pixel-wise marginal frequencies, i.e. the sampled pixel-fractions for each new stochastic class (e.g. sidewalk 2) with respect to the corresponding existing one (sidewalk)}
\label{fig:freq_m_heads_and_image2image}
\end{figure}

\begin{figure}[htbp]
\centering
\includegraphics[width=.75\textwidth]{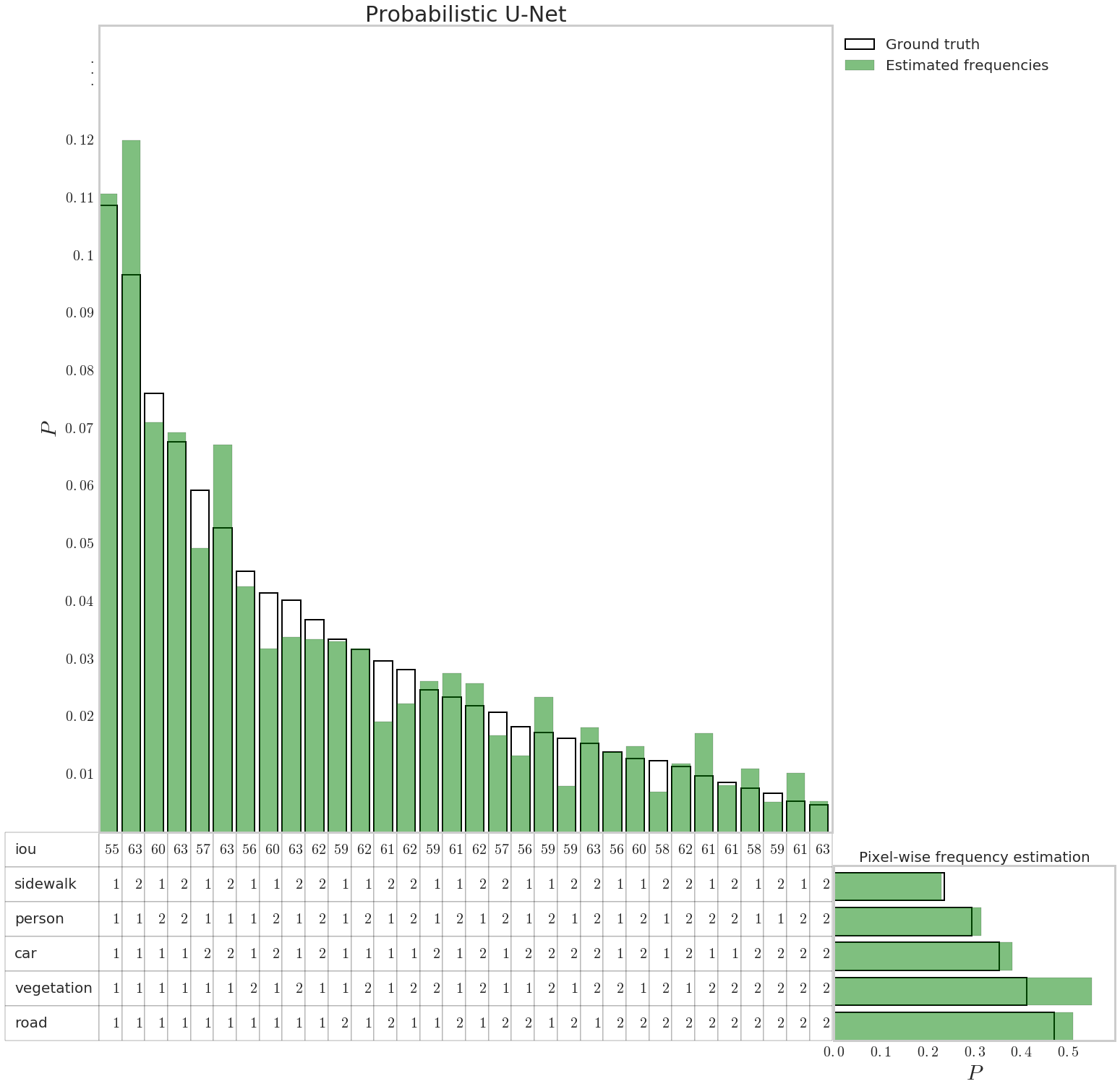}
\captionsetup{format=hang}
\caption{Reproduction of probabilities by our \probunet{}.The vertical histogram shows the mode-wise occurrence frequencies of samples in comparison to the ground-truth probability of the modes, and the horizontal histogram reports the pixel-wise marginal frequencies, i.e. the sampled pixel-fractions for each new stochastic class (e.g. sidewalk 2) with respect to the corresponding existing one (sidewalk).}
\label{fig:freq_prob_unet}
\end{figure}

In this section we analyse the frequency in which each mode of the Cityscape task is targeted by each model, and how much that varies from the ground truth distribution. 
We report the mode-wise and pixel-wise marginal occurrence frequencies of the sampled segmentation variants. 
In the mode-wise case, each sample is matched to its closest ground truth mode (using 1-IoU as the distance function). Then, the frequency of each mode is computed by counting the number of samples that most closely match that mode.
In the pixel-wise case, the marginal frequencies $p(\mathrm{predicted\, class}| \mathrm{ground}$-$\mathrm{truth\, class})$ are obtained by counting all pixels across all images and corresponding samples that show a valid pixel hypothesis given the ground-truth, normalized by the number of respective uni-modal ground-truth pixels. 
In \autoref{fig:freq_ensemble_and_dropout} we present the results for \ensemble{} and \dropoutunet, in \autoref{fig:freq_m_heads_and_image2image} we show the results for \mheads{} and \itoivae, finally in \autoref{fig:freq_prob_unet} we present the results for our approach.

\clearpage

\section{Ablation analysis}
\label{app:ablation}

In this section we explore variations in the architecture of our approach, in order to understand how each design decision affects the performance.
We have tried three variations over the original approach, these are:

\textbf{Fixing the prior}: Instead of making the prior a function of the context, here we fix it to be a standard Gaussian distribution.

\textbf{Fixing the prior, and not using the context (input image) in the posterior}: In addition to fixing the prior to be Gaussian, we also make the posterior a function of the ground truth mask only, ignoring the context. 

\textbf{Injecting the latent features at the beginning of the U-Net}: Starting from our original model, we change the position in which the latent variables are used. Specifically here we concatenate them to the context (input image) and propagate that through the U-Net.

\begin{figure}[htbp]
\centering
\includegraphics[width=.8\textwidth]{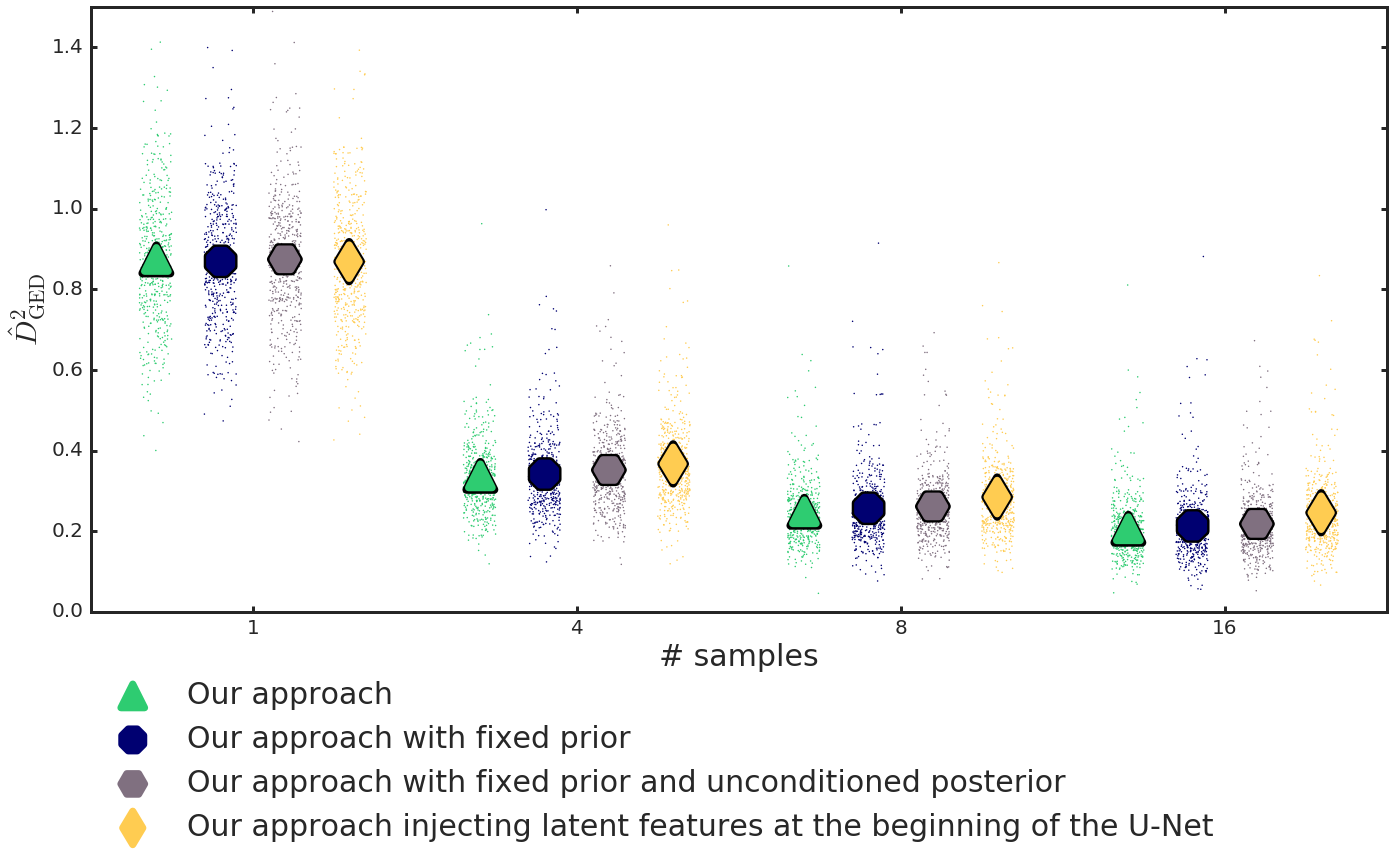}
\captionsetup{format=hang}
\caption{Ablation analysis. Comparison of architectural variations of our approach using the energy distance. Lower energy distances correspond to better agreement between predicted distributions and ground truth distribution of segmentations. The symbols that overlay the distributions of data points mark the mean performance.}
\label{fig:ablation}
\end{figure}

In \autoref{fig:ablation} we can observe that our approach is better than the other variations. As the mechanisms that induce the distributions over segmentations during sampling and training are blinded towards the context image, the performance in terms of the IoU-based energy distance decreases. In particular, our model is much better than the variation that injects latent samples at the beginning. This is a pleasant finding, given that our decision of injecting the latent variables at the end of the U-Net was motivated by efficiency reasons when sampling. Here we find that we do not lose performance by doing so, but instead observe an improved matching of the samples with the ground-truth distribution. We hypothesize that injecting the latent variables at the final stage of the pipeline makes it easier for the model to account for different segmentations given the same input. This hypothesis is supported by the slightly better performance shown by the alternative architecture when sampling only once, and how this advantage is lost, and actually reversed, when sampling several times.

\section{Predicting ground truth ambiguity from models' samples}
\label{app:ambiguity}

\begin{figure}[htbp]
\centering
\includegraphics[width=.48\textwidth]{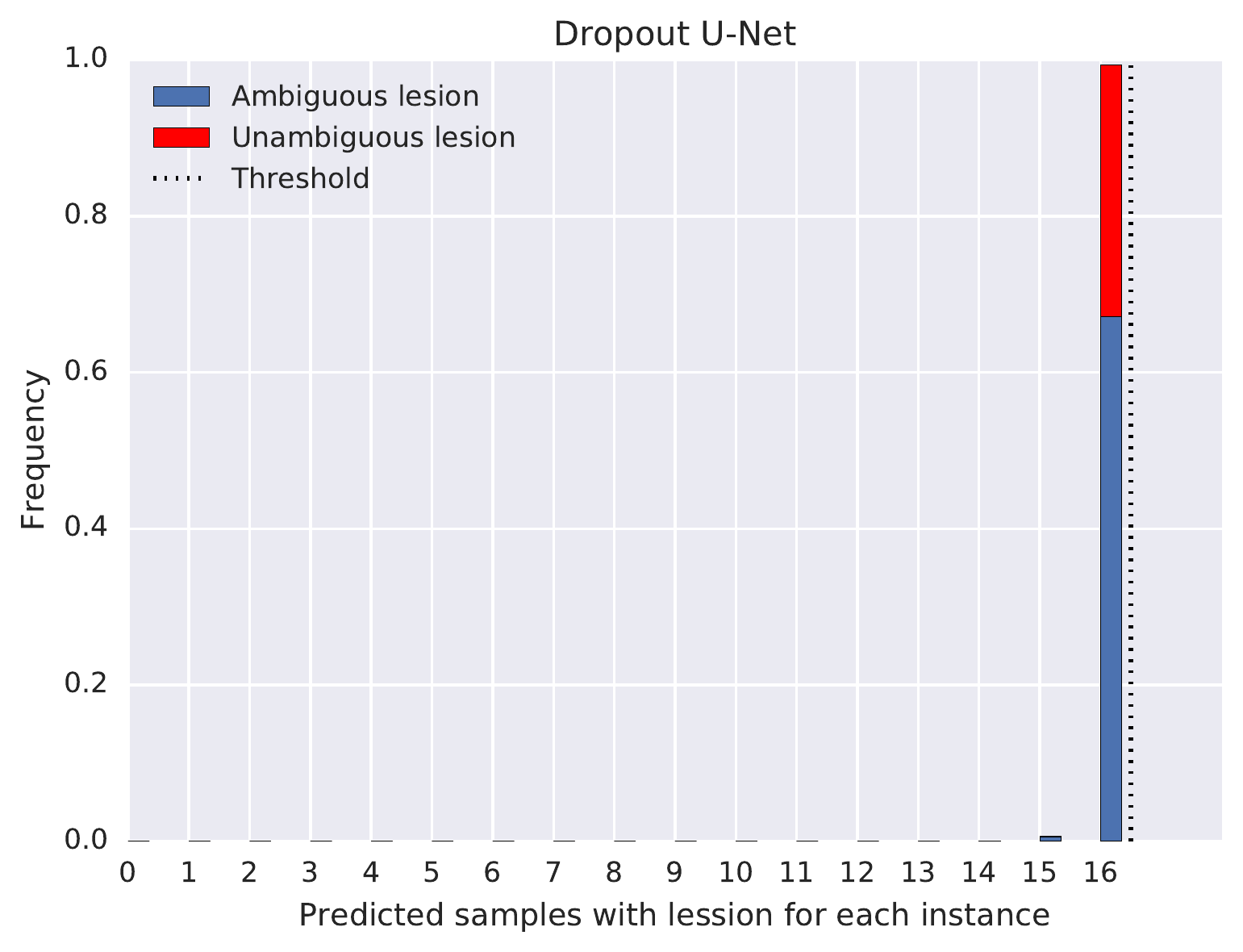}
\includegraphics[width=.48\textwidth]{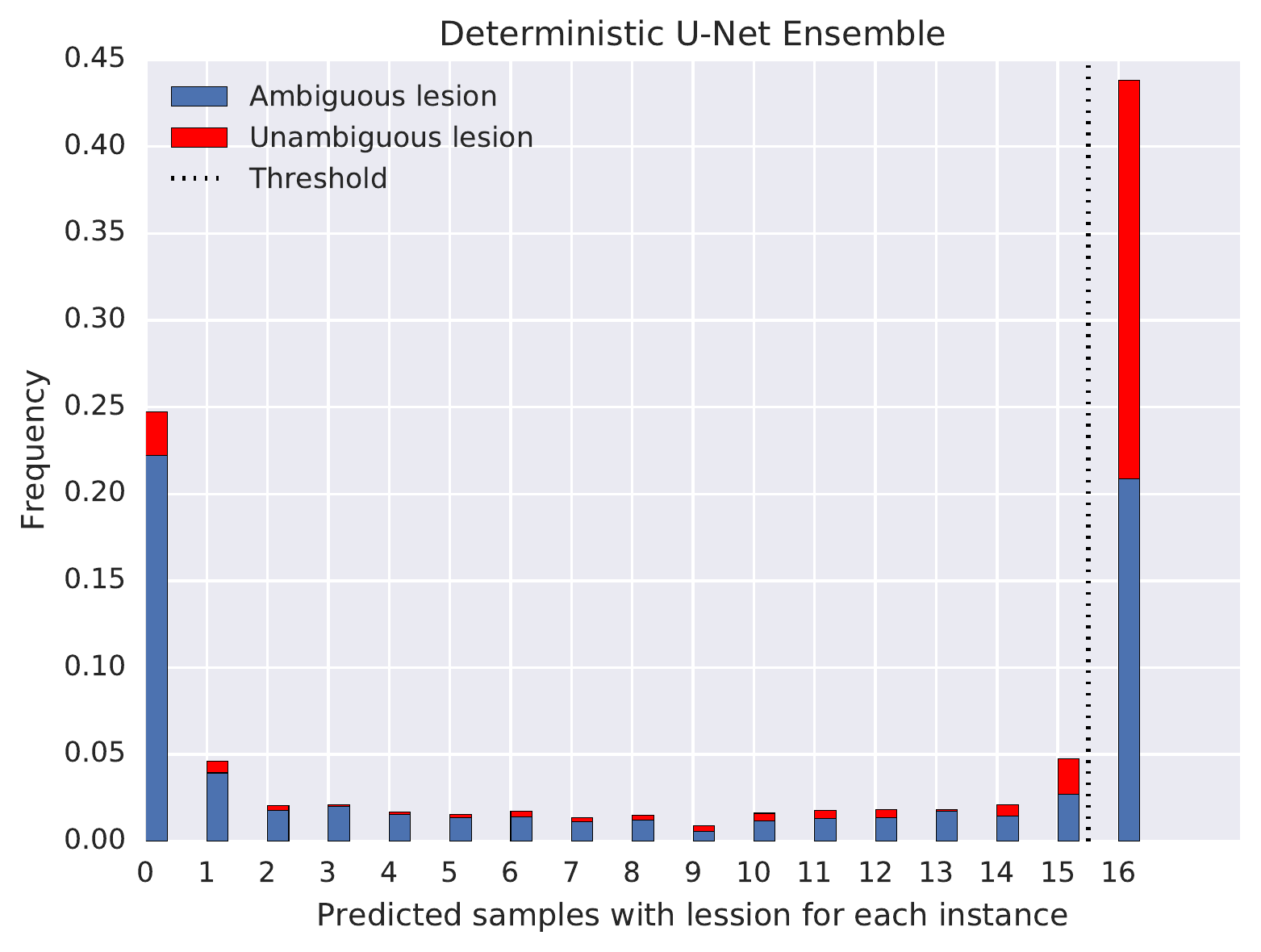}
\includegraphics[width=.48\textwidth]{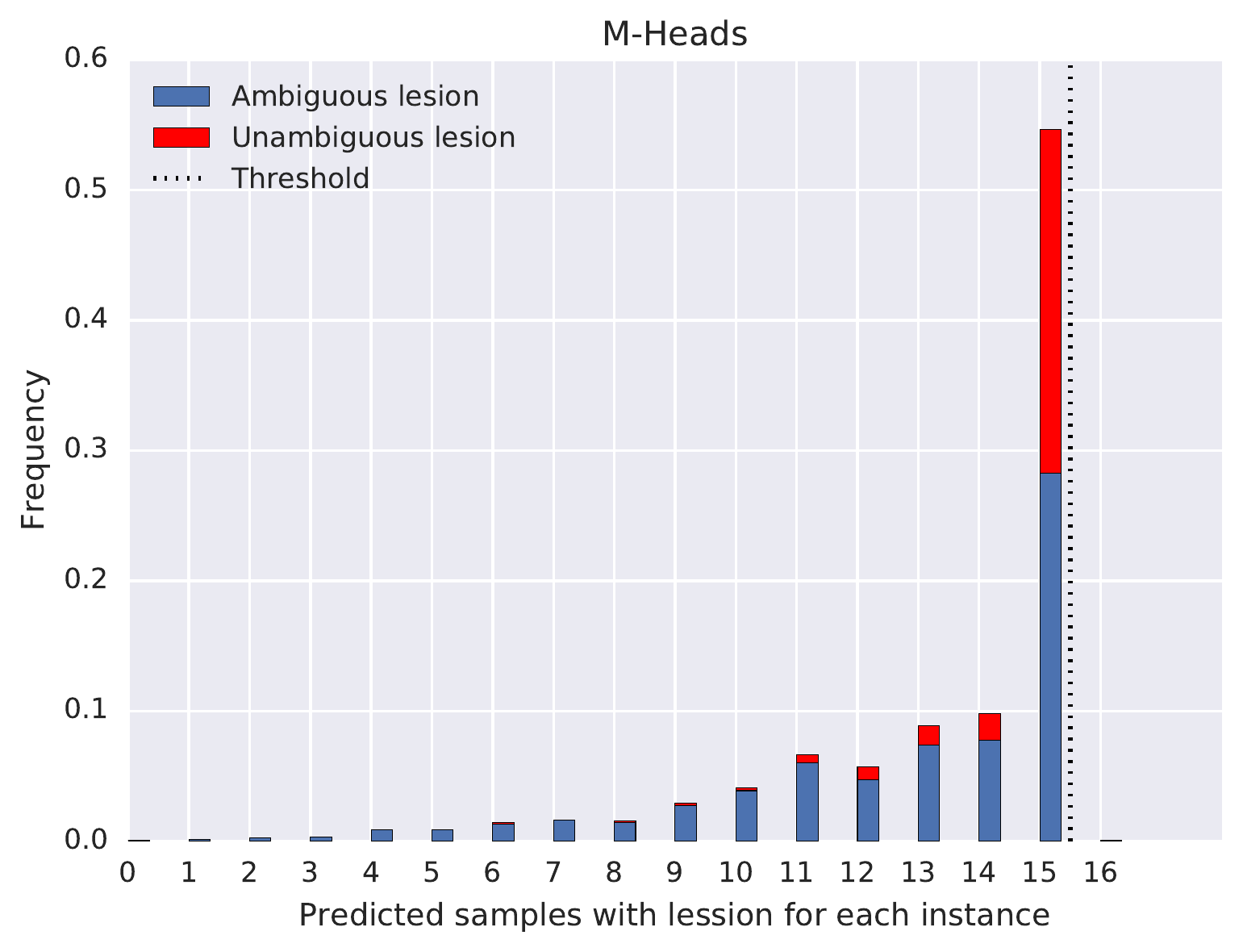}
\includegraphics[width=.48\textwidth]{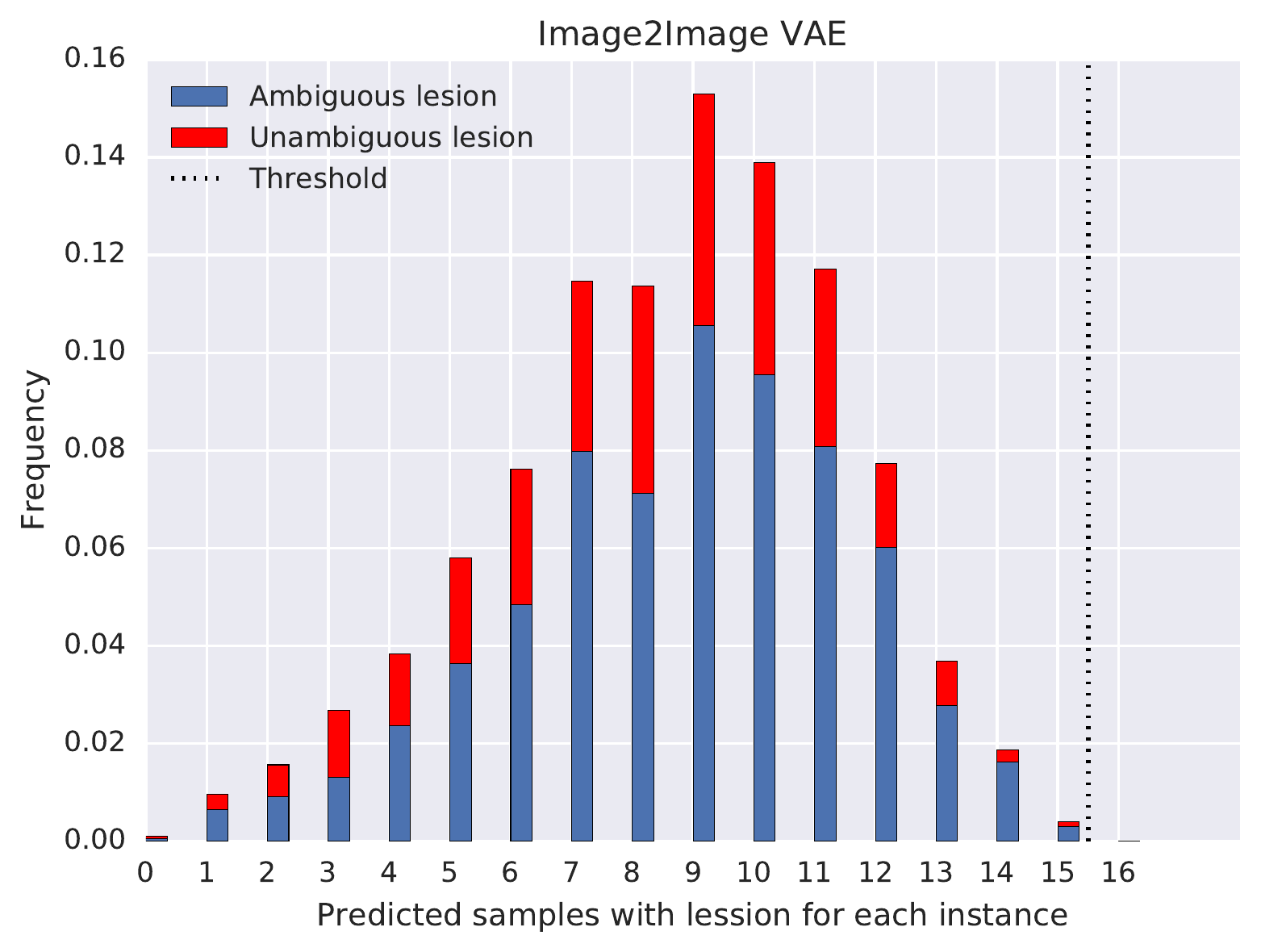}
\includegraphics[width=.48\textwidth]{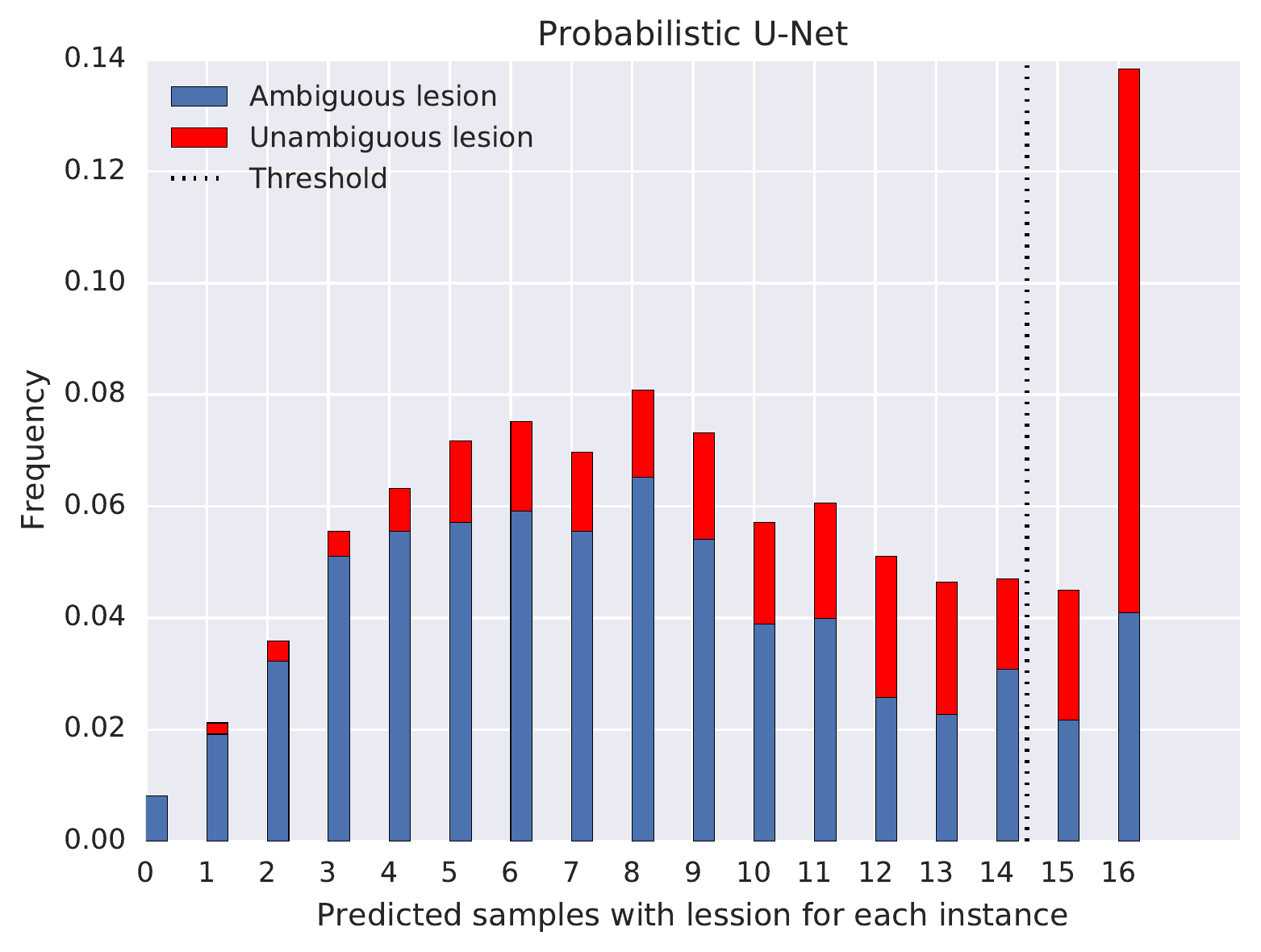}
\captionsetup{format=hang}
\caption{Histograms showing the amount of (ground truth) ambiguous and unambiguous lesions as a function of the number of times the model produces a sample with a lesion in it (out of 16 samples). Each histogram corresponds to one model.}
\label{fig:ambiguity}
\end{figure}

In this section we assess the capacity of different models trained on LIDC for distinguishing between unambiguous and ambiguous instances. Specifically we define an instance to be ambiguous if 1 or more graders disagree on the presence of abnormal tissue. To do so, for each model we draw $16$ samples per instance (as in all other experiments in the paper) and count the number of lesion presences out of the $16$. This lesion presence is binned in two histograms with $[0, 16]$ bins, one for ambiguous and one for unambiguous instances (they are shown in \autoref{fig:ambiguity}). Finally we evaluate the discriminatory power of such histograms by computing the best threshold that separates ambiguous and unambiguous instances on the validation set. We present the accuracy scores on the test set in \autoref{tab:ambiguity}, which shows the advantage that our approach has over the competitors in this regard.

\begin{table}[H]
\centering
\begin{tabular}{ c|c|c|c|c } 
 \toprule
Dropout U-Net & U-Net Ensemble & M-Heads & Image2Image VAE & Probabilistic U-Net \\
\midrule
$0.328$ & $0.699$ & $0.678$ & $0.678$ & $\bm{0.736}$ \\
\end{tabular}
\caption{Discriminative power of histograms from different models to distinguish between ambiguous and unambiguous lesions.}
\label{tab:ambiguity}
\end{table}

\section{Sampling LIDC masks using different models}
\label{app:sampling_lidc}

\begin{figure}[htbp]
\centering
\vbox{\vspace{-5mm}
\makebox[\textwidth][c]{\includegraphics[width=1.1\textwidth]{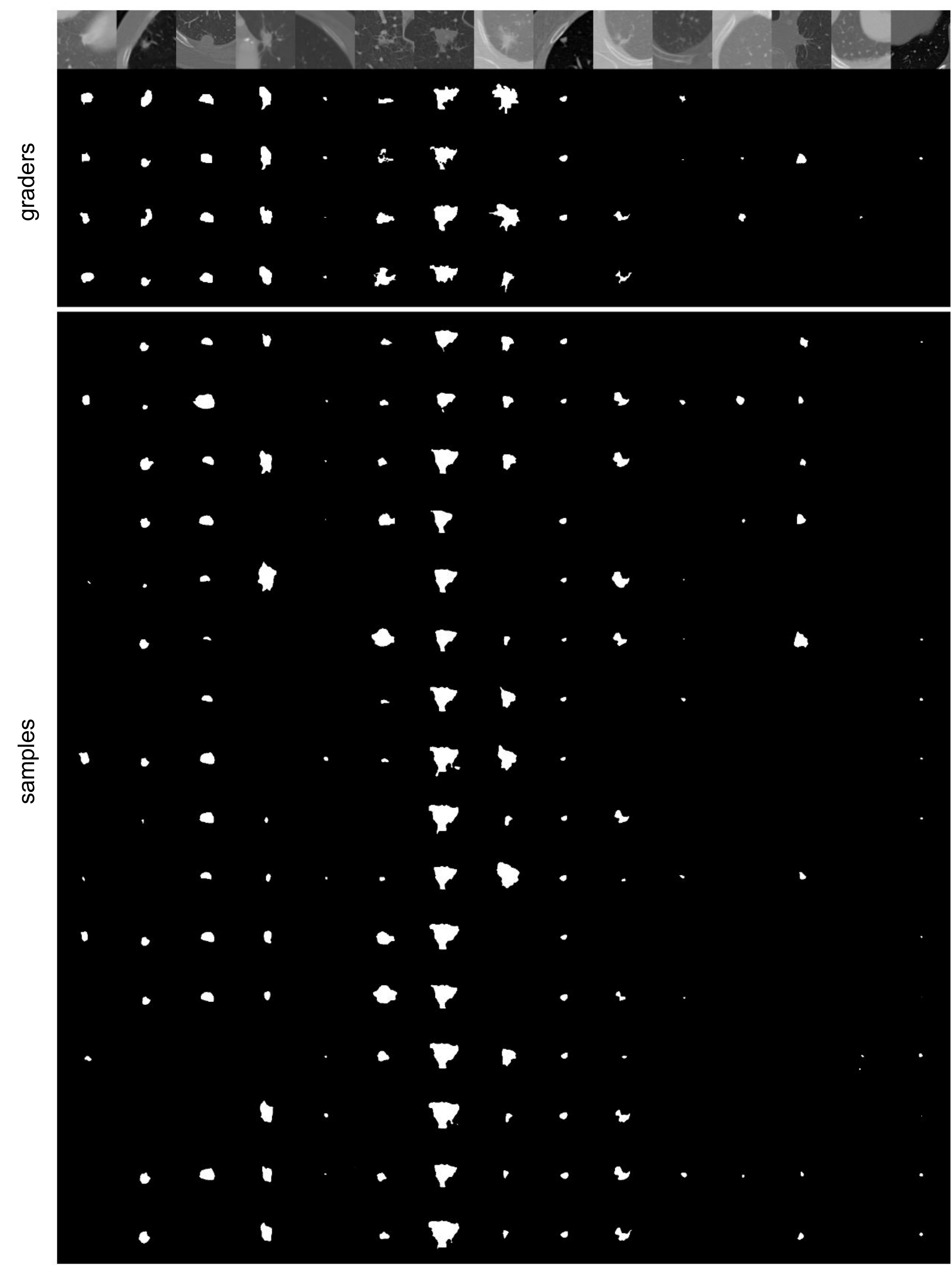}}%
\captionsetup{format=hang}
\caption{Qualitative examples from the \textbf{\probunet}. The upper panel shows LIDC test set images from 15 different subjects alongside the respective ground-truth masks by the 4 graders. The panel below gives the corresponding 16 random samples from the network.}
\label{fig:lidc_samples2}
}
\end{figure}

\begin{figure}[htbp]
\centering
\vbox{\vspace{-5mm}
\makebox[\textwidth][c]{\includegraphics[width=1.1\textwidth]{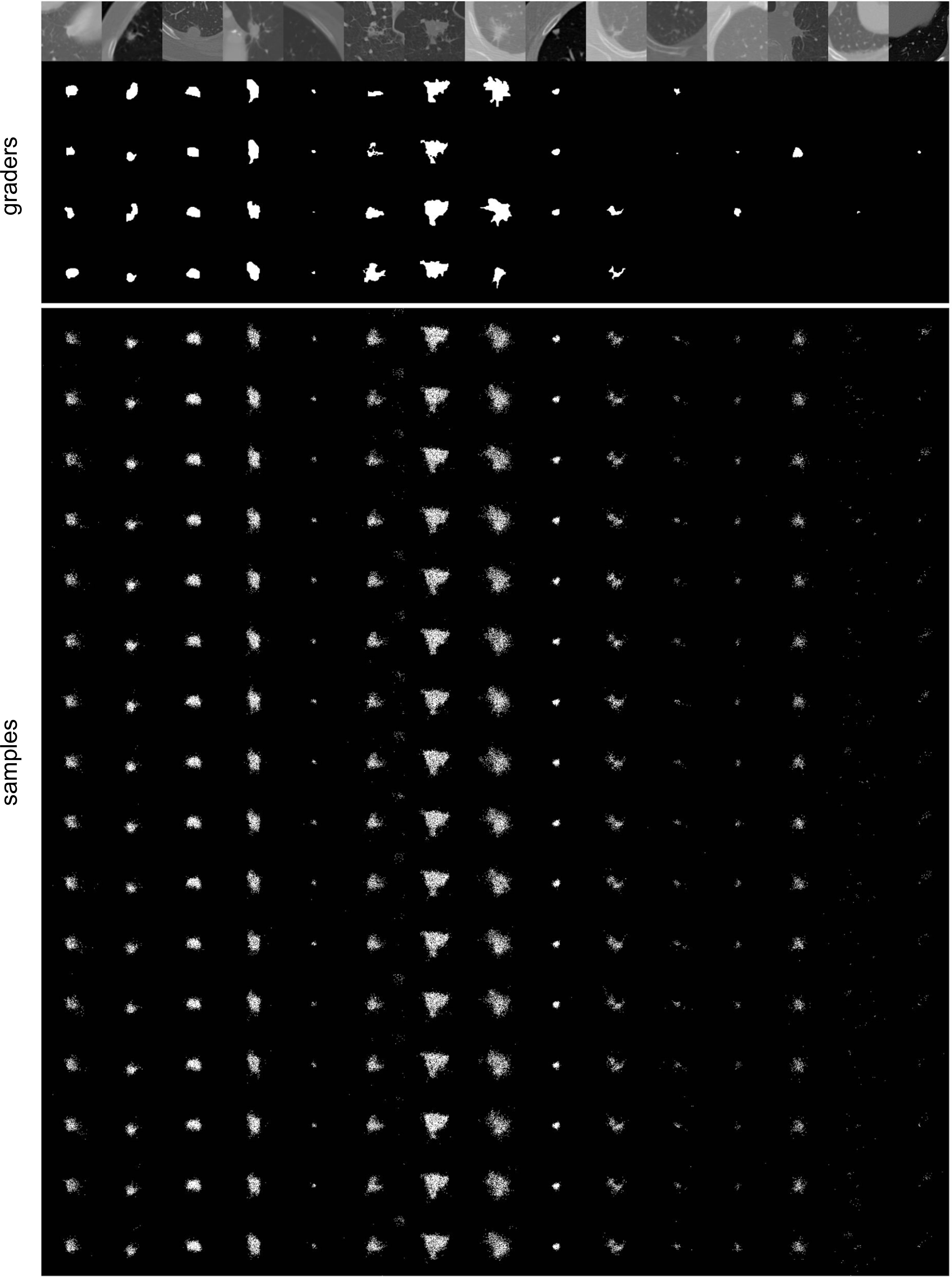}}%
\captionsetup{format=hang}
\caption{Qualitative examples from the \textbf{\dropoutunet}. Same layout as \autoref{fig:lidc_samples2}.}
\label{fig:lidc_dropout_samples}
}
\end{figure}

\begin{figure}[htbp]
\centering
\vbox{\vspace{-5mm}
\makebox[\textwidth][c]{\includegraphics[width=1.1\textwidth]{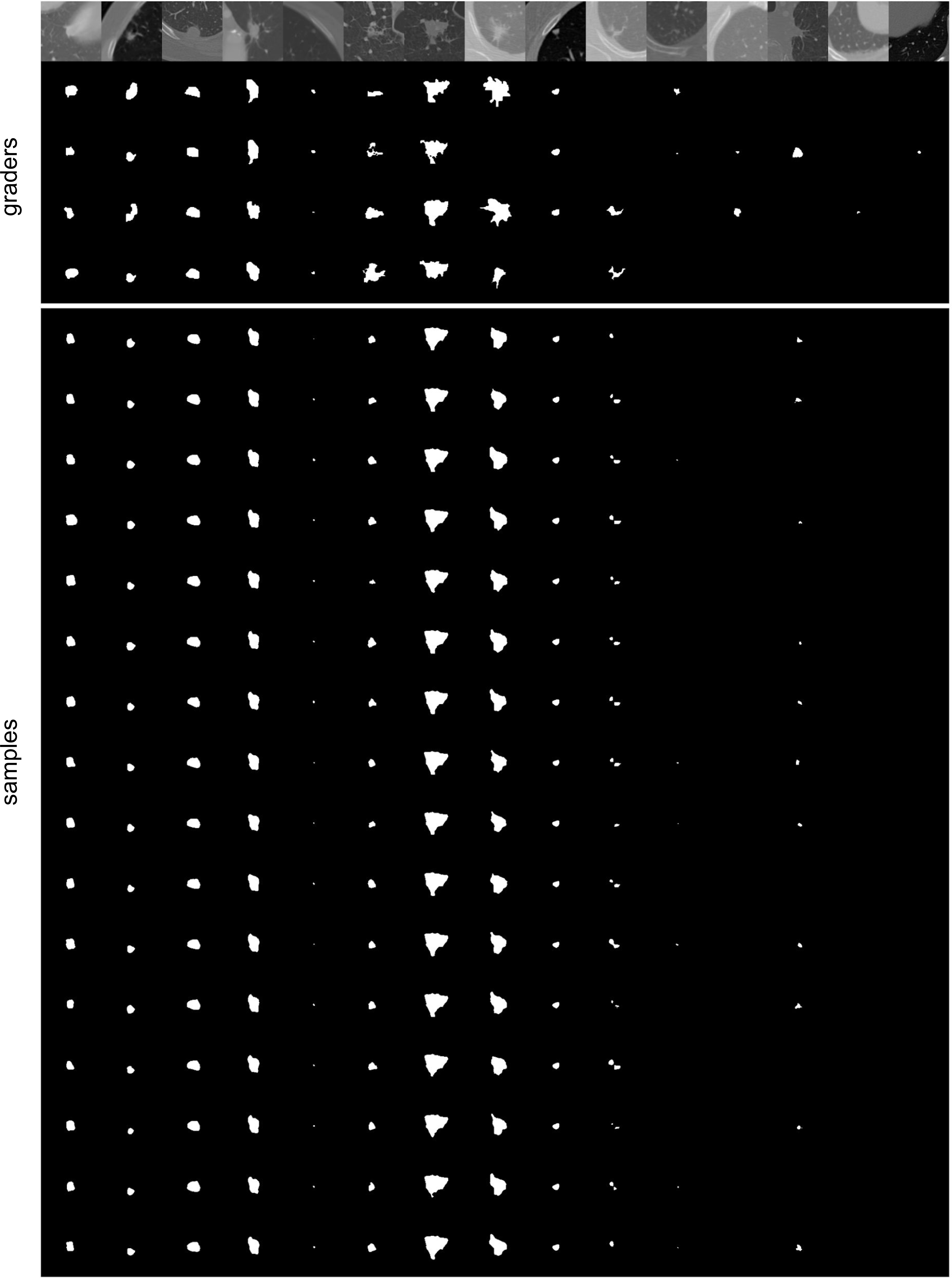}}%
\captionsetup{format=hang}
\caption{Qualitative examples from the \textbf{\ensemble}. Same layout as \autoref{fig:lidc_samples2}.}
\label{fig:lidc_ensemble_samples}
}
\end{figure}

\begin{figure}[htbp]
\centering
\vbox{\vspace{-5mm}
\makebox[\textwidth][c]{\includegraphics[width=1.1\textwidth]{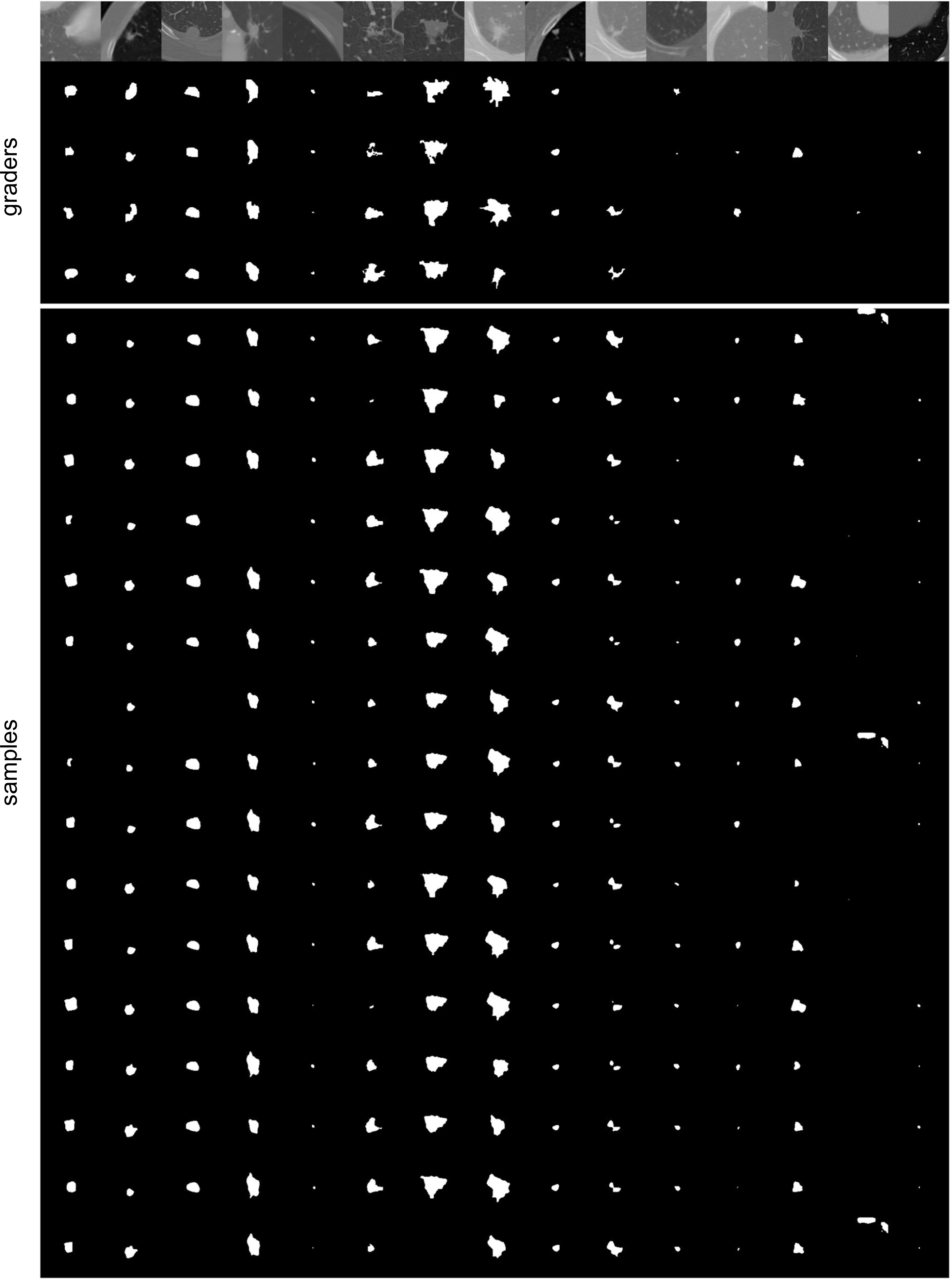}}%
\captionsetup{format=hang}
\caption{Qualitative examples from the \textbf{\mheads} (using a network with 16 heads). Same layout as \autoref{fig:lidc_samples2}.}
\label{fig:lidc_mheads_samples}
}
\end{figure}

\begin{figure}[htbp]
\centering
\vbox{\vspace{-5mm}
\makebox[\textwidth][c]{\includegraphics[width=1.1\textwidth]{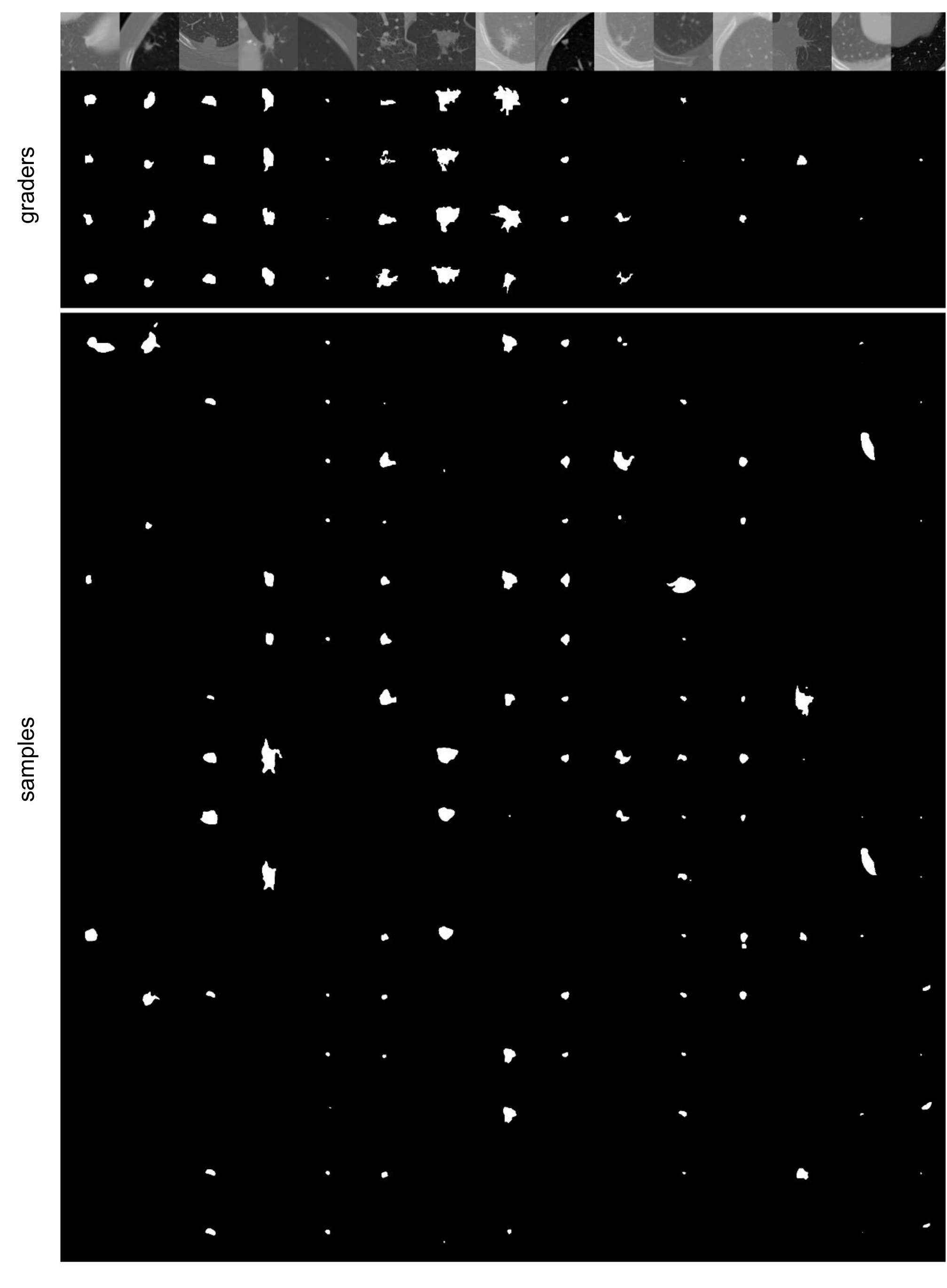}}%
\captionsetup{format=hang}
\caption{Qualitative examples from the \textbf{\itoivae}. Same layout as \autoref{fig:lidc_samples2}.}
\label{fig:lidc_image2image_samples}
}
\end{figure}

\autoref{fig:lidc_samples2}-\ref{fig:lidc_image2image_samples} show samples of our proposed model as well as all the baselines given the same input images. For reference the expert segmentations are shown in the four rows just below the images. \autoref{table:lidc_results} shows the numerical results from \autoref{fig:distance_results}a.

\begin{table}[H]
\centering
 \begin{tabular}{c | c c c c} 
 \toprule
 \# Samples & 1 & 4 & 8 & 16 \\ [0.5ex] 
 \midrule
 $\hat{D}^2_{\mathrm{GED}}$ & $0.811$ & $0.388$ & $0.321$ & $0.287$ \\  [1ex] 
 \bottomrule
\end{tabular}
\caption{Numerical (mean) results of the \probunet{} on LIDC, taken from \autoref{fig:distance_results}a.}
\label{table:lidc_results}
\end{table}

\section{Sampling Cityscapes segmentations using our model}
\label{app:sampling_cityscapes}

\begin{figure}[htbp]
\centering
\vbox{\vspace{-14mm}
\makebox[\textwidth][c]{\includegraphics[width=1.1\textwidth]{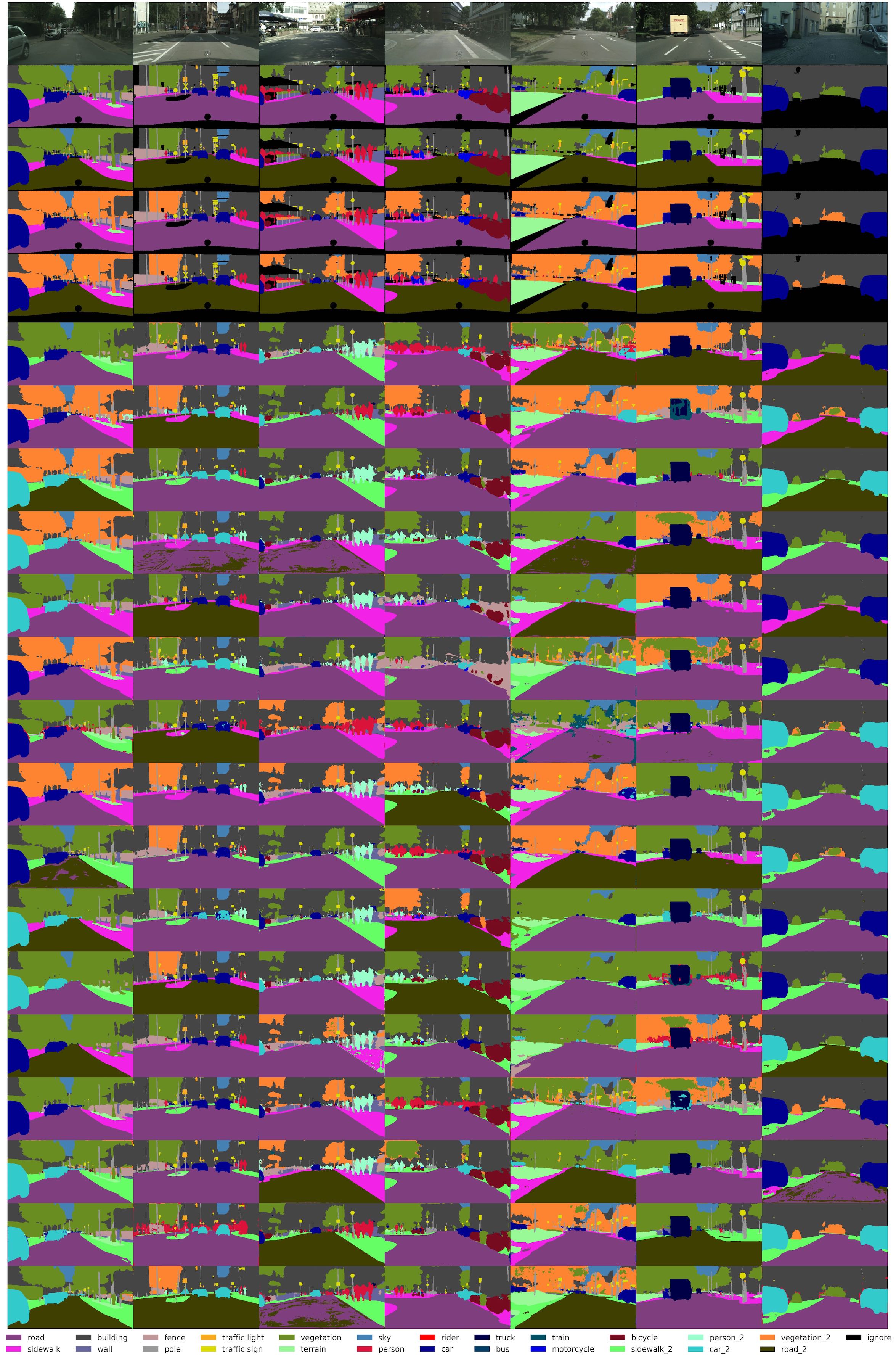}}%
\captionsetup{format=hang}
\caption{Qualitative examples from the \textbf{\probunet} on the Cityscapes task. The first row shows Cityscapes images, the following 4 rows show 4 out of the 32 ground truth modes with black pixels denoting pixels that are masked during evaluation. The remaining 16 rows show random samples of the network.}
\label{fig:cityscapes_samples}
}
\end{figure}

\autoref{fig:cityscapes_samples} shows samples of our proposed model on the Cityscapes dataset, and \autoref{table:cityscapes_results} shows the numerical results from \autoref{fig:distance_results}b, so that new approaches can be compared to those.

\begin{table}[H]
\centering
 \begin{tabular}{c | c c c c} 
 \toprule
 \# Samples & 1 & 4 & 8 & 16 \\ [0.5ex] 
 \midrule
 $\hat{D}^2_{\mathrm{GED}}$ & $0.874$ & $0.337$ & $0.248$ & $0.206$ \\  [1ex] 
 \bottomrule
\end{tabular}
\caption{Numerical (mean) results of the \probunet{} on Cityscapes, taken from \autoref{fig:distance_results}b.}
\label{table:cityscapes_results}
\end{table}

\clearpage

\section{Training details}
In this section we describe the architecture settings and training procedure for both experiments.

\subsection{Lung abnormalities segmentation}
\label{app:details_lidc}
We only use those lesions that were specified as a polygon (outline) in the XML files of the LIDC dataset, disregarding the ones that only have center of shape. That is, according to the LIDC paper we use the ones that are larger than 3mm, and filtering out the others, that are clinically less relevant \cite{armato2011lung}. We also filter out each Dicom file whose absolute value of SliceLocation differs from the absolute value of ImagePositionPatient[-1]. Finally we assume that two masks from different graders correspond to the same lesion if their tightest bounding boxes overlap.

During training image-grader pairs are drawn randomly.
We apply augmentations to the image tiles ($180 \times 180$ pixels size): random elastic deformation, rotation, shearing, scaling and a randomly translated crop that results in a tile size of $128 \times 128$ pixels. The \unet{} architecture we use is similar to \cite{Ronneberger2015} with the exception that we down- and up-sample feature maps by using bilinear interpolations. The cores of all models are identical and feature 4 down- and up-sampling operations, at each scale the blocks comprise three convolutional layers with $3 \times 3$-kernels, each followed by a ReLU-activation. 
In our model, both the prior and the posterior (as well as the posterior in \itoivae{}) nets have the same architecture as the \unet{}'s encoder path, i.e. they are made up to the same number of blocks and type of operations. Their last feature maps are global average pooled and fed into a $1 \times 1$ convolution that predicts the Gaussian distributions parameterized by mean and standard deviation.
The architecture last layers, corresponding to $f_{\text{comb.}}$, comprise the appropriate number of $1 \times 1$-kernels and are activated with a softmax. The base number of channels is 32 and is doubled or respectively halved at each down- or up-sampling transition. All individual models share this core component and for ease of comparability we let all models undergo the same training schedule: the training proceeds over $240\,\mathrm{k}$ iterations with an initial learning rate of $1e^{-4}$ that is lowered to $1e^{-6}$ in 5 steps. All weights of all models are initialized with orthogonal initialization having the gain (multiplicative factor) set to $1$, and the bias terms are initialized by sampling from a truncated normal with $\sigma=0.001$. We use a batch-size of 32, weight-decay with weight $1e^{-5}$ and optimize using the Adam optimizer with default settings \cite{kingma2014adam}.
A KL weight of $\beta=10$ with a latent space of $3$ dimensions gave best validation results for the baseline \itoivae{}, and $\beta=1$ and a $6$D latent space performed well for the \probunet{}, although the performances were alike across the hyperparameters tried on the validation set.

\subsection{Cityscapes}
\label{app:details_cityscapes}
We down-sample the Cityscapes images and label maps to a size of $256 \times 512$. Similarly to above, we apply random elastic deformation, rotation, shearing, scaling, random translation and additionally impose random color augmentations on the images during training. The \unet{} cores in this task are identical to the ones above, but process an additional feature scale (implying one additional up- and one additional down-sampling operation). The training procedure is also equivalent to the previous experiment, also using $240\,\mathrm{k}$ iterations, except that here we employ a batch-size of 16, and the initial learning rate of $1e^{-4}$ is lowered to $1e^{-5}$ in 3 steps. The Cityscapes dataset includes ignore label masks for each image with which we mask the loss during training, and the metric during evaluation. A KL weight of $\beta=1$ and 3D latents gave best validation results for the \itoivae{} and a $\beta=1$ and 6D latents performed best for the \probunet{} (although 3-5D performed similarly). 

\end{appendices}
\end{document}